\documentclass[12pt]{article}

\usepackage{setspace, graphicx, caption, float, relsize, rotating, color, verbatim, multirow,apacite}

\usepackage{mathtools} 
\mathtoolsset{showonlyrefs=true} 
\usepackage{ascmac} 

\usepackage[linesnumbered,ruled,vlined]{algorithm2e}
\usepackage{amsfonts}
\usepackage{amsmath}
\usepackage{amssymb} 
\usepackage{amsthm} 
\usepackage{authblk} 
\usepackage[english]{babel}
\usepackage{bigints} 
\usepackage{bm}
\usepackage[open,openlevel=1]{bookmark}
\usepackage{booktabs}
\usepackage{breakcites}
\usepackage{chngpage}
\usepackage{color} 
\usepackage{empheq}
\usepackage{enumerate} 
\usepackage{enumitem}
\usepackage{fancyhdr}   
\usepackage[margin=2cm]{geometry}
\usepackage{graphicx}    
\usepackage{hyperref}
\usepackage[utf8]{inputenc}
\usepackage{lastpage}   
\usepackage{listings}
\usepackage{longtable}  
\usepackage{makeidx}    
\usepackage{mathrsfs}
\usepackage{mathtools}
\usepackage{mdframed}
\usepackage{movie15}
\usepackage{multirow}
\usepackage{multirow,booktabs}
\usepackage{natbib}
\setcitestyle{authoryear} 
\usepackage{pdfpages}
\usepackage[section]{placeins}
\usepackage[document]{ragged2e}
\usepackage{sectsty}
\usepackage{times}
\usepackage{setspace}   
\usepackage{siunitx}
\usepackage{subfigure}  %
\allowdisplaybreaks
\usepackage{xcolor}
\usepackage{xmpmulti}   

\theoremstyle{definition}
\newtheorem{theorem}{Theorem}[section]

\newtheorem{lemma}{Lemma}[section]

\newtheorem{remark}{Remark}[section]

\newtheorem{condition}{C}

\newcommand{\nc}{\newcommand}

\nc{\dps}{\displaystyle}
\nc{\tr}{\text{tr}}

\title{\textbf High-Dimensional Importance-Weighted Information Criteria: Theory and Optimality}
\author[1]{Yong-Syun Cao}
\author[2]{Shinpei Imori}
\author[1]{Ching-Kang Ing}
\affil[1]{Institute of Statistics, National Tsing Hua University}
\affil[2]{Graduate School of Advanced Science and Engineering, Hiroshima University}
\date{\today}

\numberwithin{equation}{section}

\allowdisplaybreaks

\begin{document}
\setlength{\parindent}{1em}

\maketitle

\begin{abstract}
\cite{Imori202X}
proposed the importance-weighted orthogonal greedy algorithm (IWOGA) for model selection in high-dimensional misspecified regression models under covariate shift. To determine the number of IWOGA iterations, they introduced the high-dimensional importance-weighted information criterion (HDIWIC). They argued that the combined use of IWOGA and HDIWIC, IWOGA + HDIWIC, achieves an optimal trade-off between variance and squared bias, leading to optimal convergence rates in terms of conditional mean squared prediction error. In this article, we provide a theoretical justification for this claim by establishing the optimality of IWOGA + HDIWIC under a set of reasonable assumptions.

\end{abstract}
\setlength{\parindent}{0.3in}
\section{Introduction}
Various methods for high-dimensional model selection have been developed in recent years to address situations where the training and test data come from different distributions. When both input and output variables are available in the source (training) and target (test) domains but the target sample size is small, estimates based solely on the target data often suffer from high variance. To improve accuracy, auxiliary estimates from the source domain can be incorporated, along with bias correction to account for domain differences. This transfer learning strategy facilitates more reliable estimation under limited target information 
(see, for example,  \cite{li2020transfer}, \cite{bastani2021predicting}, and \cite{tian2022transfer}).

However, when test outputs (i.e., target responses) are unavailable, estimation or bias correction involving both domains becomes infeasible, as only inputs (covariates) are observed in the test set. In such cases, importance weighting is commonly used to reweight the source data, allowing for effective parameter estimation and model selection tailored to the target domain.
This study focuses on the latter setting. While allowing input distributions to differ between domains, we assume the conditional distribution of the response given covariates remains the same--an assumption known as covariate shift (see \cite{shimodaira2000jspi} and \cite{sugiyama2012machine}).

In a recent paper, \cite{Imori202X} 
proposed reweighting the training samples using trimmed density ratios between the test and training covariates. The resulting weights were incorporated into a modified version of the Orthogonal Greedy Algorithm (OGA), termed the Importance-Weighted OGA (IWOGA), for high-dimensional regression model selection under covariate shift. They showed that IWOGA achieves a fast convergence rate in test prediction error
as the number of iterations increases, even under model misspecification.
However, the optimal number of iterations--necessary for achieving the optimal convergence rate--depends on the sparsity level of the regression coefficients, which is typically unknown in practice. To address this,
\cite{Imori202X}
proposed a data-driven method, the high-dimensional importance-weighted information criterion (HDIWIC), to determine the appropriate number of IWOGA iterations. This paper aims to show that the combination of IWOGA and HDIWIC, referred to as IWOGA+HDIWIC, achieves the optimal convergence rate in high-dimensional regression under covariate shift and model misspecification, without requiring prior knowledge of the sparsity level.

	
The remainder of the paper is organized as follows. Section \ref{sec:sec2.1} reviews the preliminaries of
\cite{Imori202X}, including notation and assumptions. Section \ref{sec:sec2.2} presents the main result, showing that IWOGA+HDIWIC effectively achieves optimal performance. In Section \ref{sec:sec3}, we return to the conventional OGA and demonstrate that when the full high-dimensional regression model is correctly specified, the standard HDIC, introduced in \cite{ing2020model}, can be used to determine the optimal number of iterations under covariate shift, particularly when the difference between the training and test input distributions is moderate, as formalized in condition \eqref{eq:eq92}. 
All technical details are provided in Section \ref{sec5}.

\setlength{\parindent}{0pt}
\section{Convergence rate of IWOGA + HDIWIC}
\label{sec:sec2}
	\subsection{
     Framework of IWOGA under Covariate Shift: Notation and Assumptions}
        \label{sec:sec2.1}
    \setlength{\parindent}{1em}
 Let $y\in\mathbb{R}$ denote the response variable and $\bm{x} = (x_1,\dots,x_{p_n})^\top$ denote the covariates, on which we add superscript "$tr$" or "$te$" if they are from training or testing distribution. Consider a linear regression model for the test data, 
	\begin{align}
    \label{ing25000}
		y^{te}
			 = \alpha + \bm{\beta}^\top \bm{x}^{te} + \varepsilon^{te}
			 = y^{te}(\bm{x}^{te}) + \varepsilon^{te}, 
	\end{align}
	where $\alpha$ and $\bm{\beta}$ are the coefficients of the best linear predictor under the full model $J_F=\{1, \ldots, p_n\}$. 
    For a subset $J \subset J_F$ with $J \neq \emptyset$,
     $\alpha(J)$ and $\bm{\beta}(J)$ denote 
     the corresponding coefficients of the best linear predictor of $y^{te}$ restricted to model $J$. 
     Throughout this paper, we assume model $J (\subset J_F)$ is not empty.
     We allow model misspecification in the sense that
     $E[y^{te}|\bm{x}^{te}] \neq y^{te}(\bm{x}^{te})$.
     Let $f^{tr}(\bm{x})$ and $f^{te}(\bm{x})$ denote the training and test input densities, and
   $f^{tr}(y|\boldsymbol{x})$ and $f^{te}(y|\boldsymbol{x})$
the conditional densities of $y$ given $\boldsymbol{x}$
in the training and test data.   
Although $f^{tr}(\bm{x})$ and $f^{te}(\bm{x})$ may not coincide, the covariate shift assumption requires that
$f^{tr}(y|\boldsymbol{x})=f^{te}(y|\boldsymbol{x})$,
yielding
$E[y^{tr}|\boldsymbol{x}^{tr}=\bm{x}]
=E[y^{te}|\boldsymbol{x}^{te}=\bm{x}]$,
which we denote by
$y(\boldsymbol{x})$ in what follows.
Note that when $J_F$ is correctly specified,
\begin{align}
\label{ing250018}
y(\bm{x})= \alpha + \bm{\beta}^\top \bm{x}.
\end{align}
Moreover,
we assume in the rest of the paper that $f^{tr}(\bm{x})$ and $f^{te}(\bm{x})$ have a common support,
and
$0<w(\bm{x})=f^{te}(\boldsymbol{x})/f^{tr}(\bm{x}) <\infty$
for any $\bm{x}$ in the support.

    Let  $(y^{tr}_t, \bm{x}^{tr}_t)$, $t = 1, \ldots, n$,
    denote the observed training data.
    Define the {\it importance} by $w(\bm{x})$. 
   Rather than directly using
    $w(\bm{x}^{tr}_t)$ to reweight 
    the $t$th training sample,  
    \cite{Imori202X} considered 
    a truncated version:
    \begin{align}
    \label{ing25001}
    v(\bm{x}^{tr}_t) = \min\{ w(\bm{x}^{tr}_t), b_n \},
    \end{align}
    where $b_n$ is a trimming 
    parameter
    introduced to stabilize the potentially large values of 
    $w(\bm{x}^{tr}_t)$.
    The final weight applied in IWOGA is then defined as
    \begin{align}
     \label{ing25002}
    w_t = \frac{v(\bm{x}^{tr}_t)}{ n^{-1}\sum\limits_{t=1}^n v(\bm{x}^{tr}_t) }.
    \end{align}
    For further details on IWOGA, refer to Section 2 of
    \cite{Imori202X}.

Let $\hat{J}^{te}_m$ denote the model selected by IWOGA
at the $m$th iteration.
For notational simplicity, 
we will omit the “{\it te}” superscript in $\hat{J}^{te}_m$
throughout the paper.
Using $\hat{J}_m$,
$y^{te}(\bm{x}^{te})$ in \eqref{ing25000}
can be predicted by
 $\hat{y}^{te}( \bm{x}^{te}|\hat{J}_m)$,
 where
\begin{align}
\label{ing25004}
\hat{y}^{te}(\boldsymbol{x}|J)=\hat{\alpha}(J)+\hat{\boldsymbol{\beta}}(J)^\top\boldsymbol{x}_J
\end{align}
is the weighted least squares predictor
based on model $J$.
Note that
for $\boldsymbol{v} \in \mathbb{R}^{p_n}$,
$\boldsymbol{v}_J \in \mathbb{R}^{\sharp{J}}$ denotes its subvector corresponding to the index set $J$, where $\sharp{J}$ denotes the cardinality of $J$, and
$\hat{\alpha}(J) \in \mathbb{R}$ and $\hat{\boldsymbol{\beta}}(J) \in \mathbb{R}^{\sharp{J}}$
are the weighted least squares
estimates of $\alpha(J)$ and $\boldsymbol{\beta}(J)$; 
see \eqref{ing250013} for details.
The performance of 
$\hat{y}^{te}(\boldsymbol{x}^{te}|\hat{J}_m)$
is measured by
the modified conditional prediction error
(MCPE),
\begin{align}
\label{ing25005}
\begin{split}
&E_n[(y^{te}(\boldsymbol{x}^{te})-
\hat{y}^{te}(\boldsymbol{x}^{te}|\hat{J}_m)
)^2]\\
=&
E[(y^{te}(\boldsymbol{x}^{te})-\hat{y}^{te}(\boldsymbol{x}^{te}|\hat{J}_m)
)^2|
(y^{tr}_i, \bm{x}^{tr}_i), i=1 \ldots,n
].
\end{split}
\end{align}
To investigate the convergence rate 
of the MCPE as $m$ increases,
\cite{Imori202X} imposed the following conditions.



        \begin{condition}
            \label{cond:cond1}
            There exists a constant $\gamma\in(0,1)$ such that $\log p_n = O(n^{\gamma})$, and $p_n \rightarrow \infty$ as $n \rightarrow \infty$. 
        \end{condition}

        \begin{condition}
            \label{cond:cond2}
            There exist constants $\theta_0>0$ and $\eta\in(0,2]$ such that
    		\begin{align*}
    			&\underset{n\rightarrow\infty}{\text{lim sup }} E[\exp(\theta_0|\varepsilon^{tr}|^\eta)] < \infty,~~\underset{n\rightarrow\infty}{\text{lim sup }} E[\exp(\theta_0|\varepsilon^{te}|^\eta)] < \infty,  \\
    			&\underset{n\rightarrow\infty}{\text{lim sup }} \max\limits_{0\leq i,j\leq p_n} E[\exp(\theta_0|x_i^{te}x_j^{te}|)] < \infty, 
    		\end{align*}
            where $\varepsilon^{tr} = y^{tr} - y(\bm{x}^{tr})$ and  $x_0^{te} = 1$. 
        \end{condition}
        
        \begin{condition}
            \label{cond:cond3}
            There exists $q > 0$ such that
            \begin{align*}
                \underset{n\rightarrow\infty}{\text{lim sup }} E[w(\bm{x}^{te})^q] < \infty. 
            \end{align*}
        \end{condition}
        
        \begin{condition}
            \label{cond:cond4}
            $\{ (\varepsilon^{tr}_t, \bm{x}^{tr}_t) \}_{t=1}^{n}$ are independent and identically distributed, 
            where $\varepsilon^{tr}_t = y^{tr}_t - y(\bm{x}^{tr}_t)$. 
        \end{condition}
        
        \begin{condition}
            \label{cond:cond6}
            $\sup\limits_{n\geq1} (|\alpha| + \Vert \bm{\beta} \Vert_2) < \infty$ and there exist constants $\xi \geq 0$ and $C_\xi > 0$ such that for all $J\subset J_F$,
            \begin{align*}
                \Vert \bm{\beta}_J \Vert_1 \leq C_\xi \Vert \bm{\beta}_J \Vert_2^{\frac{2\xi}{2\xi+1}}.
            \end{align*}
        \end{condition}
        
       In addition,
       \cite{Imori202X}
       assumed that there exist constants $C^{te}_{\bm{\Sigma}}>0$ and $C^{te}_{\lambda+} > 0$ such that
        \begin{align}
            \max\limits_{\substack{1\leq \sharp{J}\leq K_n \\ j\notin J}} \Vert ( \bm{\Sigma}^{te}_{J,J} )^{-1} \bm{\sigma}^{te}_{J,j} \Vert_1 \leq C^{te}_{\bm{\Sigma}}, \label{eq:eq51} \\
            \lambda_{\min}( \bm{\Sigma}^{te} ) \geq C^{te}_{\lambda+}, \label{eq:eq52}
        \end{align}
        where $\bm{\Sigma}^{te} \in \mathbb{R}^{p \times p}$ is the covariance matrix of $\bm{x}^{te}$, $\bm{\Sigma}^{te}_{J, J}$ is the submatrix of $\bm{\Sigma}^{te}$ corresponding to $J$, $\bm{\sigma}^{te}_{J,j} = (\bm{\sigma}^{te}_j)_J$, and $\bm{\sigma}^{te}_j$ is the $j$th column of $\bm{\Sigma}^{te}$. The counterparts
        of $\bm{\Sigma}^{te}$, $\bm{\Sigma}^{te}_{J, J}$, $\bm{\sigma}^{te}_j$, and $\bm{\sigma}^{te}_{J, j}$
        for the training data are denoted by $\bm{\Sigma}^{tr}$, $\bm{\Sigma}^{tr}_{J, J}$, $\bm{\sigma}^{tr}_j$, and $\bm{\sigma}^{tr}_{J, j}$, respectively.
        
        Let $c_n = (\log p_n / n)^{1/2}$,
    		\begin{align}
    			d_n = d_n(q) = 
    				\begin{cases}
    					c_n^{\frac{2q}{1+q}} (\log n)^{\frac{1}{\eta} + \frac{1}{2}}, &~0 < q \leq 1, \\
    					c_n, &~ q>1,
    				\end{cases} \label{eq:eq53}
    		\end{align}
		and $K_n$
        be a prescribed upper bound for the number of iterations
        of IWOGA.
 Under the above assumptions,
        \cite{Imori202X} established the following result regarding the convergence rate of 
        \eqref{ing25005}.

        \begin{theorem}[Theorem 3 of \cite{Imori202X}]
            \label{thm:thm1}
            Assume (C\ref{cond:cond1}) -- (C\ref{cond:cond6}), \eqref{eq:eq51} and \eqref{eq:eq52}. Suppose that
            for some sufficiently small $M_k > 0$, 
            \begin{align}
            \label{ing250011}
            K_n = M_k d_n^{-1},
            \end{align}
            and
            for some $M_w > 0$ and $M_\eta \geq 1/\eta + 1/2$,
            $b_n$ (see  \eqref{ing25001}) satisfies
    		\begin{align}
    			b_n = b_n(q) = 
    				\begin{cases}
    					M_w c_n^{-\frac{2}{1+q}}, &~0 < q \leq 1, \\
    					M_w c_n^{-1} (\log n)^{-M_\eta}, &~ q>1.
    				\end{cases} \label{eq:eq54}
    		\end{align}
Then,
    		\begin{align}
\label{ingfeb65}
\max_{1 \leq m \leq K_n} \frac{E_n[(y^{te}(\boldsymbol{x}^{te})-\hat{y}^{te}(\boldsymbol{x}^{te} | \hat{J}_m))^2]}{L^{(\xi)}_n(m)} = O_p(1),
\end{align}
where $L^{(\xi)}_n(m)=m^{-(1+2\xi)}+md_n^2$.
Moreover,
            \begin{align}
            \label{eq:eq1033}
    			\max\limits_{1\leq m \leq K_n} \frac{ E_n[ \varepsilon^{te}( \bm{x}^{te} | \hat{J}_m )^2 ] }{ m^{-(1+2\xi)} + d_n^{\frac{1+2\xi}{1+\xi}} } = O_p(1), 
    		\end{align}
    		where  $\varepsilon^{te}(\bm{x}|J) = y^{te}(\bm{x}) - y^{te}( \bm{x}|J)$
and $y^{te}( \bm{x}^{te}|J)=
 \alpha(J)+ \boldsymbol{\beta}(J)^\top \boldsymbol{x}^{te}_J$
 is the  population counterpart of
$\hat{y}^{te}(\boldsymbol{x}^{te}|J)$.
        \end{theorem}
Equation \eqref{ingfeb65}
indicates that
the best choice for $m$
is $d_n^{-\frac{1}{1+\xi}}$, which yields
a convergence rate of
$d_n^{\frac{1+2\xi}{1+\xi}}$.
However, this choice 
is not adaptive, as it depends on
$\xi$, which is typically unknown in practice.
 In the next subsection, we demonstrate that this optimal rate can still be achieved when 
$m$ is selected via HDIWIC.
 A key step in the proof involves 
\eqref{eq:eq1033},
whose proof, along with that of
\eqref{ingfeb65},
is provided in Appendix C of
 \cite{Imori202X}.
 
If $w(\cdot)$ is unknown, the weight $w_t$ in \eqref{ing25001}
    can be replaced by
\begin{align}
     \label{ing25003}
    \hat{w}_t = \frac{\hat{v}(\bm{x}^{tr}_t)}{ n^{-1}\sum\limits_{t=1}^n \hat{v}(\bm{x}^{tr}_t) },
    \end{align}
    where $\hat{v}(\bm{x}^{tr}_t)$ is defined as
    $v(\bm{x}^{tr}_t)$ with 
    $w(\cdot)$ superseded by
    a suitable estimate
    $\hat{w}(\cdot)$.
Proposition 5 of \cite{Imori202X}
shows that \eqref{ingfeb65} and \eqref{eq:eq1033}
in Theorem \ref{thm:thm1}
continue to hold 
when \eqref{ing25002} is replaced 
by \eqref{ing25003}, provided that
$\hat{v}(x^{tr}_t)$
in \eqref{ing25003} satisfies the following conditions:
    		\begin{align}
    				P\left( \max\limits_{0 \leq i \leq j \leq p_n} \left| \frac{1}{n} \sum\limits_{t=1}^n (v(x^{tr}_t) - \hat{v}(x^{tr}_t))x^{tr}_{ti} x^{tr}_{tj} \right| \geq C d_n \right) = o(1),
 \label{eq:eq55}
    \end{align}
and
\begin{align}
                    \max\limits_{0 \leq i \leq p_n} \left| \frac{1}{n} \sum\limits_{t=1}^n (v(x^{tr}_t) - \hat{v}(x^{tr}_t)) x^{tr}_{ti} \varepsilon^{tr}_{t} \right| = O_{p}(d_{n}),
    		 \label{eq:eq550}
    		\end{align}
where $C>0$ is some constant.
Section 3.2 of
\cite{Imori202X} introduces 
an estimate $\hat{w}(\cdot)$
of the
importance weight $w(\cdot)$ under Gaussian assumptions
and
shows that the resulting $\hat{v}(\bm{x}_t)$
satisfies \eqref{eq:eq55} and \eqref{eq:eq550}
under suitable regularity conditions.
For more details, see 
Appendix E of
\cite{Imori202X}. 
          
%
%
   
    \subsection{Main results}
        \label{sec:sec2.2}
We start by introducing HDIWIC in the setting where
$w(\cdot)$ is known.
Let $\bm{y}^{tr} = (y^{tr}_1, \ldots, y^{tr}_n)^\top$, $\bm{X}^{tr} = (\bm{x}^{tr}_1, \ldots, \bm{x}^{tr}_n)^\top$, and $\bm{W} = {\rm diag}\{w_1, \ldots, w_n\}$. 
    Define $\hat{\mu}^{te}_y = \bm{1}_n^\top \bm{W} \bm{y}^{tr}/n$, $\hat{\bm{\mu}}^{te} = (\bm{X}^{tr})^\top \bm{W} \bm{1}_n/n$, $\hat{\bm{\Sigma}}^{te} = (\bm{X}^{tr})^\top\bm{W}^{1/2}\bm{P}_{\bm{W}}^\perp\bm{W}^{1/2}\bm{X}^{tr}/n$, and $\hat{\bm{s}}^{te} = (\bm{X}^{tr})^\top\bm{W}^{1/2}\bm{P}_{\bm{W}}^\perp\bm{W}^{1/2}\bm{y}^{tr}/n$, where $\bm{P}_{\bm{W}}^\perp = \bm{I}_n - \bm{P}_{\bm{W}}$ and $\bm{P}_{\bm{W}} = n^{-1}\bm{W}^{1/2}\bm{1}_n\bm{1}_n^\top \bm{W}^{1/2}$, which is the orthogonal projection matrix onto the column space spanned by $\bm{W}^{1/2}\bm{1}_n$. 
For $J \subset J_F$ with $1 \leq \sharp J \leq K_n$, let
        \begin{align}
            (\hat{\sigma}^{te}(J))^2 &= \frac{1}{n}\sum\limits_{t=1}^n w_t (y^{tr}_t - \hat{y}^{te}(\bm{x}_t^{tr}|J))^2, \label{eq:eq1}
        \end{align}
        where $\hat{y}^{te}(\bm{x}|J)$
        is defined in \eqref{ing25004}, with
        \begin{align}
        \label{ing250013}
        \hat{\alpha}(J) = \hat{\mu}^{te}_y - \hat{\bm{\beta}}(J)^\top \hat{\bm{\mu}}^{te}_J, \,\,\hat{\bm{\beta}}(J) = (\hat{\bm{\Sigma}}^{te}_{J,J})^{-1}\hat{\bm{s}}^{te}_J.
        \end{align}
    Note that 
    \begin{align}
        (\hat{\sigma}^{te}(J))^2 &= \frac{1}{n}(\bm{y}^{tr})^\top \bm{W}^{1/2} (\bm{I}_n -\bm{P}_{\bm{W}} - \bm{P}(J))\bm{W}^{1/2} \bm{y}^{tr}, 
    \end{align}
    where $\bm{P}(J) = \bm{P}_{\bm{W}}^\perp\bm{W}^{1/2}\bm{X}^{tr}_{\cdot,J}((\bm{X}^{tr}_{\cdot,J})^\top\bm{W}^{1/2}\bm{P}_{\bm{W}}^\perp\bm{W}^{1/2}\bm{X}^{tr}_{\cdot,J})^{-1}(\bm{X}^{tr}_{\cdot,J})^\top\bm{W}^{1/2}\bm{P}_{\bm{W}}^\perp$, which is the orthogonal projection matrix onto the column space spanned by $\bm{P}_{\bm{W}}^\perp\bm{W}^{1/2}\bm{X}^{tr}_{\cdot,J}$, and $\bm{A}_{\cdot, J} = (a_{ij})_{1 \leq i \leq p_n, j \in J}$ for a matrix $\bm{A} = (a_{ij})_{1 \leq i,j \leq p_n} \in \mathbb{R}^{p_n \times p_n}$, that is, $\bm{X}^{tr}_{\cdot,J} = (\bm{x}_{1, J}, \ldots, \bm{x}_{n, J})^\top$. 
    Then, the HDIWIC value is defined as
    \begin{align}
        \label{ing250010}
        \text{HDIWIC}(J)= (1 + s_a (\sharp{J}+1) d_n^2) (\hat{\sigma}^{te}(J))^2, 
    \end{align}
    where $s_a$ is a positive constant. 
    Recall that $\hat{J}_k$ denotes the model selected by IWOGA at the $k$th iteration. Define
        \begin{align}
                \begin{split}
                    \hat{k}_n
                        &= \hat{k}
                        = \underset{1\leq k \leq K_n}{\text{arg min}}~\text{HDIWIC}( \hat{J}_k ).
                \label{eq:eq2}
        \end{split}
        \end{align}
        %
    The following theorem assesses the performance of 
    $\hat{y}^{te}( \bm{x}^{te} | \hat{J}_{\hat{k}} )$ in predicting $y^{te}(\bm{x}^{te})$.
    
        \begin{theorem}
            \label{thm:thm2}
            Suppose that the assumptions of Theorem \ref{thm:thm1}
            hold, and 
            $s_a$ in \eqref{ing250010}
            is sufficiently large.
            Then,
            \begin{align}
            \label{ing250016}
                \frac{ E_n [y^{te}(\bm{x}^{te}) - \hat{y}^{te}( \bm{x}^{te} | \hat{J}_{\hat{k}} )]^2 }{ d_n^{\frac{1+2\xi}{1+\xi}} } = O_p(1). 
            \end{align}
	\end{theorem}
\noindent
    In light of the discussion following
 Theorem \ref{thm:thm1},
 \eqref{ing250016} implies that
 $\hat{y}^{te}( \bm{x}^{te} | \hat{J}_{\hat{k}} )$
 achieves the optimal rate, $d_n^{\frac{1+2\xi}{1+\xi}}$,
 without requiring knowledge of the value of $\xi$.
 
Next, we consider the case where the
$w_t$'s are unavailable.
In this case, they are replaced by
$\hat{w}_t$'s computed using
$\hat{v}(\bm{x}^{tr}_t)$'s.
Under this replacement, let $\check{J}_m$
denote the model selected by
IWOGA at the $m$th iteration.
Moreover, define
        \begin{align}
        \label{ing250012}
            \begin{split}
                (\check{\sigma}^{te}(J))^2
                    &= \frac{1}{n}\sum\limits_{t=1}^n \hat{w}_t 
                        (
                            y^{tr}_t - \check{y}^{te}( \bm{x}_t^{tr}|J )
                        )^2,
            \end{split} 
        \end{align}
where $\check{y}^{te}( \bm{x} |J)=
\check{\alpha}(J)+
    \check{\bm{\beta}}(J)^{\top}
    \bm{x}_J$, with
    $\check{\alpha}(J)$
    and $ \check{\bm{\beta}}(J)$
    being defined in the same way as
     $\hat{\alpha}(J)$
    and $ \hat{\bm{\beta}}(J)$, respectively,
    except that the $w_t$'s are replaced by $\hat{w}_t$'s,
 \begin{align}
  \label{ing250015}
~\text{HDIWIC}_s(J)= \left(
                                    1 + s_a (\sharp{J}+1) d_n^2
                                \right) (\check{\sigma}^{te}(J))^2, 
  \end{align}
and 
        \begin{align}
                    \check{k}_n
                        &= \check{k}
                        = \underset{1\leq k \leq K_n}{\text{arg min}}~\text{HDIWIC}_s( \check{J}_k).
        \end{align}
        %
    %
    The next theorem examines the performance of
    $\check{y}^{te}( \bm{x}^{te} | \check{J}_{\check{k}} )$
    in predicting $y^{te}(\bm{x}^{te})$.

        \begin{theorem}
            \label{thm:thm3}
            Assume (C\ref{cond:cond1})--(C\ref{cond:cond6}), \eqref{eq:eq51}, \eqref{eq:eq52}, \eqref{eq:eq55},
            \begin{align}
                &P\left(\max\limits_{0 \leq i \leq p_n} \left| \frac{1}{n} \sum\limits_{t=1}^n (v(x^{tr}_t) - \hat{v}(x^{tr}_t)) x^{tr}_{ti} \varepsilon^{tr}_{t} \right| \geq Cd_n \right) = o(1), \label{eq:eq133}
            \end{align}
            and
            \begin{align}
                 \frac{1}{n} \sum\limits_{t=1}^n 
 (v(x^{tr}_t) - \hat{v}(x^{tr}_t))
 (\varepsilon^{tr}_{t})^2 = o_p(1).
                \label{imr20Mar18_1}
            \end{align}
            Then, for sufficiently large $s_a$ in \eqref{ing250015},
    		\begin{align}
             \label{ing250017}
    			\frac{ E_n  [y^{te}(\bm{x}^{te}) - 
                \check{y}^{te}( \bm{x}^{te} | \check{J}_{\check{k}} )
                ] ^2 }{ d_n^{\frac{1+2\xi}{1+\xi}} } = O_p(1). 
    		\end{align}
        \end{theorem}
\noindent
Theorem \ref{thm:thm2}
indicates that
the optimal rate
$d_n^{\frac{1+2\xi}{1+\xi}}$
can be achieved by 
IWOGA+HDIWIC$_{s}$, without requiring knowledge of 
either 
$\xi$ or the importance weight function
$w(\cdot)$.
Note that 
$E_n[\cdot]$
in \eqref{ing250017}
is understood as the conditional expectation
given the training data 
and (possibly) additional information used to construct
$\hat{v}(\cdot)$, both of which are assumed to be independent of
$\boldsymbol{x}^{te}$.
Finally, we note that
\eqref{eq:eq133} and
\eqref{imr20Mar18_1} can be justified in a manner similar to the justification of \eqref{eq:eq55}
and \eqref{eq:eq550}.

\section{Convergence rate of OGA + HDIC}
    \label{sec:sec3}
When there is no covariate shift,
\cite{ing2020model}
showed that the optimal iteration number for OGA (which is IWOGA with $w_t=1$) can be selected by minimizing HDIC over the sequence of nested models generated by the algorithm. 
With a slight abuse of notation, we continue to denote by
$\hat{J}_m$ the model selected by OGA at the $m$th iteration.
Under covariate shift, Section 3.3 of \cite{Imori202X} established the convergence rate of the conditional prediction error (CPE) for the least-squares predictor 
$\hat{y}^{tr}(\bm{x}^{te}|\hat{J}_m)$,
$E_n[(y(\boldsymbol{x}^{te})-\hat{y}^{tr}(\boldsymbol{x}^{te}|\hat{J}_m))^2]$, assuming that
$J_F$ is correctly specified
and without requiring condition
(C\ref{cond:cond3}). 
The predictor takes the form
\begin{align}
\label{ing250019}
\hat{y}^{tr}(\bm{x}^{te}|J)=
 \hat{\alpha}^{tr}(J)+\hat{\boldsymbol{\beta}}^{tr}(J)^\top\boldsymbol{x}_J,
\end{align}
where
$\hat{\alpha}^{tr}(J)$ and $\hat{\bm{\beta}}^{tr}(J)$ are defined as $\hat{\alpha}(J)$ and $\hat{\bm{\beta}}(J)$ with $w_t = 1$,
and serve as estimators of 
$\alpha^{tr}(J)$ and $\bm{\beta}^{tr}(J)$,
the coefficients of the best linear predictor of $y^{tr}$ based on 1 and $\bm{x}^{tr}_J$.
Specifically, 
besides  (C\ref{cond:cond1}), (C\ref{cond:cond2}), (C\ref{cond:cond4}), and (C\ref{cond:cond6}),
they assumed:
    \begin{condition}
        \label{cond:cond7}
        $J_F$ is correctly specified, namely, \eqref{ing250018} holds.
    \end{condition}
    
    \begin{condition}
        \label{cond:cond8}
        There exists a constant $\theta_0 > 0$ such that
    	\begin{align}
    		\underset{n\rightarrow\infty}{\text{lim sup}} \max\limits_{0 \leq i \leq j \leq p_n} E[ \exp( \theta_0 | x^{tr}_i x^{tr}_j | ) ] < \infty, 
    	\end{align}
        where $x_0^{tr} = 1$. 
    \end{condition}
\noindent
They also imposed analogues of  
   \eqref{eq:eq51} and \eqref{eq:eq52}:
    \begin{align}
        \max\limits_{\substack{1\leq \sharp{J}\leq K_n \\ j\notin J}} \Vert ( \bm{\Sigma}^{tr}_{J,J} )^{-1} \bm{\sigma}^{tr}_{J,j} \Vert_1 < C^{tr}_{\bm{\Sigma}}, \label{eq:eq61} \\
        \lambda_{\min}( \bm{\Sigma}^{tr}) > C^{tr}_{\lambda+}, \label{eq:eq62}
    \end{align}
 where $C^{tr}_{\bm{\Sigma}}$ and $C^{tr}_{\lambda+}$
 are some positive constants.
 Consequently, their results 
 concerning the performance of
 $\hat{y}^{tr}(\bm{x}^{te}|\hat{J}_m)$ 
  are summarized in the following theorem.
    \begin{theorem}[Theorem 6 of \cite{Imori202X}]
        \label{thm:thm4}
        Let $K_n = M_K c_n^{-1}$, where $M_K > 0$ is a sufficiently small constant.
\begin{itemize}
    \item [(i)]
        Assume that (C\ref{cond:cond1}), (C\ref{cond:cond2}), (C\ref{cond:cond4})--(C\ref{cond:cond8}), \eqref{eq:eq61}, and \eqref{eq:eq62} hold, then
    \begin{align}
\label{ingmar90}
\max_{1 \leq m \leq K_n}\frac{E_n[(y(\boldsymbol{x}^{te})-\hat{y}^{tr}(\boldsymbol{x}^{te}|\hat{J}_m))^2]}{m^{-2\xi} + m^2c_n^2} = O_p(1).
\end{align}
          
        \item [(ii)]
        Furthermore, if there exists a positive constant $C_{\text{diff}} > 0$ such that, for all $n$,
        \begin{align}
            \Vert \bm{\mu}^{te} - \bm{\mu}^{tr} \Vert_2 \leq C_{\text{diff}},~\Vert \bm{\Sigma}^{te} - \bm{\Sigma}^{tr} \Vert_2 \leq C_{\text{diff}}, \label{eq:eq92}
        \end{align}
        then,
        \begin{align}
\label{ingmar91}
\max_{1 \leq m \leq K_n}\frac{E_n[(y(\boldsymbol{x}^{te})-\hat{y}^{tr}(\boldsymbol{x}^{te}|\hat{J}_m))^2]}{m^{-(1+2\xi)} + mc_n^2} = O_p(1).
\end{align}
    \end{itemize}
\end{theorem}
 Equation \eqref{ingmar90} indicates that the CPE 
of $\hat{y}^{tr}(\boldsymbol{x}^{te}|\hat{J}_m^{tr})$ is mainly determined by the sum of a variance term, $m^2c_n^2$, and a squared bias term, $m^{-2\xi}$.
To
balance these two terms,
one may select
\begin{align}
\label{ing250020}
m=c_n^{-\frac{1}{1+\xi}},
\end{align}
leading
to an error bound,
\begin{align}
\label{ingmarr92}
\left(\frac{\log p_n}{n}\right)^{\frac{\xi}{1+\xi}}=
c_n^{\frac{2\xi}{1+\xi}}.
\end{align}
Furthermore, if the distributions of the
training and test inputs do not have notably different first and second moments, as specified in \eqref{eq:eq92},
then \eqref{ingmar91} reveals that the same choice
$m=c_n^{-\frac{1}{1+\xi}}$ balances the
variance and squared bias, yielding
a faster rate of convergence of
$c_n^{\frac{1+2\xi}{1+\xi}}$.
This rate is known to be the minimax-optimal rate for parameter estimation
under the sparsity condition
(C\ref{cond:cond6})
without covariate shift.
For more details, see \cite{Imori202X}. 

Unfortunately,
\eqref{ing250020}, which depends on the unknown $\xi$,
is not directly applicable in practice.
To address this issue,
 we propose selecting 
$m$ in a data-driven fashion
using HDIC. For a model $J$, its HDIC value is
defined as
    \begin{align}
            \text{HDIC}(J)= \left(
                            1 + s_a (\sharp{J}+1) c_n^2
                        \right) (\hat{\sigma}^{tr}(J))^2, \label{eq:eq6663}
    \end{align}
where $s_a$ is a positive constant 
and
\begin{align}
        (\hat{\sigma}^{tr}(J))^2
                    = \frac{1}{n}\sum\limits_{t=1}^n  (y^{tr}_t - \hat{y}^{tr}( \bm{x}_t^{tr}|J ))^2 
    \end{align}
is the residual mean squared error
of model $J$ 
when fitted to the training data.
In addition, define
    \begin{align}
        \hat{k}_n
            = \hat{k}
            = \underset{1\leq k \leq K_n}{\text{arg min}}~\text{HDIC}( \hat{J}_k ), \label{eq:eq63}
    \end{align}
noting that the symbol $\hat{k}_n$
was previously used in 
\eqref{eq:eq2} with a slightly different meaning. We retain this notation here for simplicity as long as no confusion arises.
The performance of
$\hat{y}^{tr}( \bm{x}^{te} | \hat{J}_{\hat{k}})$
in predicting $y(\bm{x}^{te})$
is summarized in the following theorem.
    \begin{theorem}
        \label{thm:thm5}
        Assume the same assumptions as in Theorem
        \ref{thm:thm4} (i) hold.
        Then, for sufficiently large $s_a$
        in \eqref{eq:eq6663}, 
	\begin{align}
            \frac{ E_n[(y(\bm{x}^{te}) - \hat{y}^{tr}( \bm{x}^{te} | \hat{J}_{\hat{k}}))^2] }{ c_n^{\frac{2\xi}{1+\xi}} } = O_p(1). \label{eq:eq95}
	\end{align}
        Furthermore, if \eqref{eq:eq92} holds, then
	\begin{align}
            \frac{ E_n[(y(\bm{x}^{te}) - \hat{y}^{tr}( \bm{x}^{te} | \hat{J}_{\hat{k}}))^2] }{ c_n^{\frac{1+2\xi}{1+\xi}} } = O_p(1). \label{eq:eq103}
	\end{align}
    \end{theorem}
As shown in \eqref{eq:eq95} and
\eqref{eq:eq103},
$\hat{k}$ in \eqref{eq:eq63}
can be used instead of the unknown
$c_n^{-\frac{1}{1+\xi}}$ to
help OGA achieve the optimal convergence rate under 
condition (C\ref{cond:cond7})
in the presence of covariate shift.

\section{Proofs of Theorems \ref{thm:thm2}, \ref{thm:thm3}, and \ref{thm:thm5}}
 \label{sec5}
The proof of Theorem \ref{thm:thm2} involves a delicate combination of
Theorem 3 from \cite{Imori202X} and
Theorem 3.1 from \cite{ing2020model}. Similarly, the proof of Theorem \ref{thm:thm3} requires a nontrivial integration of Proposition 5 of \cite{Imori202X}
with Theorem 3.1 from \cite{ing2020model}. Theorem \ref{thm:thm5} is proved by carefully combining Theorem 6 of \cite{Imori202X}
with the same result from \cite{ing2020model}. In all cases, aligning the assumptions and techniques across these sources requires technical care and subtle reasoning.
Section \ref{sec:sec5}
presents several lemmas that support the proofs of Theorems \ref{thm:thm2} and \ref{thm:thm3}
in Section \ref{sec:sec6},
and Theorem
\ref{thm:thm5} in Section \ref{sec:sec7}.

    \subsection{Some basic lemmas}
    \label{sec:sec5}
    \begin{lemma}
        \label{lem:lem1}
        (Lemmas 7 and 9 of \cite{Imori202X}).
        Assume (C\ref{cond:cond1}) - (C\ref{cond:cond4}), then there exist positive numbers $C_1$ and $M$ such that for all large $n$,
        \begin{align}
            &P\left( \max\limits_{0 \leq i \leq j \leq p_n} \left| \frac{1}{n} \sum\limits_{t=1}^n v(x^{tr}_t) x^{tr}_{ti} x^{tr}_{tj} - E[x^{te}_{i} x^{te}_J] \right| \geq M d_n \right) \leq 4(p_n+1)^2 \exp(- C_1M \log p_n),   \\
            &P\left( \max\limits_{0 \leq i \leq p_n} \left| \frac{1}{n} \sum\limits_{t=1}^n v(x^{tr}_t) x^{tr}_{ti} \varepsilon^{tr}_{t} \right| \geq M d_n \right) \leq 4 (p_n+1) \exp(- C_1M \log p_n) + O(n^{-1}). 
        \end{align}
    \end{lemma}

    By Lemma \ref{lem:lem1},  \cite{Imori202X} further developed 
    in Remark 11 that
    \begin{align}
        \label{ing0105}
        \begin{split}
            & P\left( \max_{0\leq i, j\leq p_n} \left|\frac{1}{n}\sum_{t=1}^nw_tx^{tr}_{ti}x^{tr}_{tj} - E[x^{te}_ix^{te}_j] \right| > \overline{B} d_n \right) = o(1),\\
            & P\left( \max_{0\leq i \leq p_n} \left|\frac{1}{n}\sum_{t=1}^n w_t x^{tr}_{ti}\varepsilon^{tr}_t \right| > \overline{B} d_n \right) = o(1).
        \end{split}
    \end{align}
    Throughout this paper, $\overline{B}$ denotes a generic positive constant independent of $n$, whose value may vary from place to place.
    
    By (C\ref{cond:cond2}) and \eqref{ing0105}, it follows that 
    \begin{align}
        \label{imr250430-2}
        &P\left(\Vert \hat{\bm{\mu}}^{te} \Vert_\infty > \overline{B} \right) = o(1), \\
        \label{imr250430}
        &P\left( \max_{1\leq i \leq j\leq p_n} |\hat{\sigma}^{te}_{ij} - \sigma^{te}_{ij} | > \overline{B} d_n \right) = o(1),
    \end{align}
    where $\sigma^{te}_{ij}$ and $\hat{\sigma}^{te}_{ij}$ are the $(i, j)$ element of $\bm{\Sigma}^{te}$ and $\hat{\bm{\Sigma}}^{te}$, respectively, that is, 
    \begin{align}
        \hat{\sigma}^{te}_{ij} = \frac{1}{n}\sum_{t=1}^nw_t(x^{tr}_{ti} - \hat{\mu}^{te}_i)(x^{tr}_{tj} - \hat{\mu}^{te}_j), ~~ \hat{\mu}^{te}_i = \frac{1}{n} \sum\limits_{t=1}^n w_t x^{tr}_{ti}. 
    \end{align}

    \begin{lemma}
        \label{lem:lem2} (Lemma 17 of \cite{Imori202X}).
        Assume (C\ref{cond:cond6}) and \eqref{eq:eq52}. Then all $J \subset J_F$,
        \begin{align}
            \left\Vert \bm{\beta}_{J^c} \right\Vert_1 \leq C  E[ \varepsilon^{te}(\bm{x}^{te}|J)^2 ] ^{\frac{\xi}{1+2\xi}}. 
        \end{align}
        Here and throughout, $C$ denotes a generic
        positive constant whose value may vary from one occurrence to another.
    \end{lemma}

    The following Lemma can be obtained from the proof of Lemma 13 of \cite{Imori202X}: 
    \begin{lemma}
        \label{lem:lem3}
        (Lemma 13 of \cite{Imori202X}).
        Under the same conditions of Theorem \ref{thm:thm1}, it holds that for all $J \subset J_F$ with $1 \leq \sharp{J} \leq K_n$, 
        \begin{align}
            P\left(\max\limits_{1 \leq \sharp{J} \leq K_n} \Vert ( \hat{\bm{\Sigma}}_{J,J}^{te} )^{-1} \Vert_2 > \overline{B} \right) = o(1). 
        \end{align}
    \end{lemma}
    

    %
    Let $\bm{\beta}^\star(J) \in\mathbb{R}^{p_n}$ be the augmented vector of $\bm{\beta}(J) \in \mathbb{R}^{\sharp{J}}$, that is, $(\bm{\beta}^\star(J))_J = \bm{\beta}(J)$ and $(\bm{\beta}^\star(J))_{J^c} = \bm{0}_{p_n - \sharp{J}}$. 
    
    \begin{lemma}
        \label{lem:lem4}
        Assume (C\ref{cond:cond6}), \eqref{eq:eq51}, and \eqref{eq:eq52}, then for all $J \subset J_F$, 
        \begin{align}
            \Vert\bm{\beta} - \bm{\beta}^\star(J)\Vert_1 \leq 
            C E[\varepsilon^{te}(\bm{x}^{te}|J)^2]^{\frac{\xi}{1+2\xi}}. 
        \end{align}
    \end{lemma}

    {\sc proof}. 
    By Lemma \ref{lem:lem2} and \eqref{eq:eq51}, we have        
    \begin{align}
        \Vert\bm{\beta} - \bm{\beta}^\star(J)\Vert_1 &= \left\Vert \bm{\beta}_J - \bm{\beta} (J) \right\Vert_1 + \left\Vert \bm{\beta}_{J^c} \right\Vert_1  \\
        &\leq \left( C^{te}_{\bm{\Sigma}} + 1 \right) \left\Vert \bm{\beta}_{J^c} \right\Vert_1 
        \label{imr250506-5}\\
        &\leq C \left( C^{te}_{\bm{\Sigma}} + 1 \right) E[\varepsilon^{te}(\bm{x}^{te}|J)^2]^{\frac{\xi}{1+2\xi}}, 
    \end{align}
   which leads to the desired conclusion. 

    
    We now define  
    \begin{align}
        \bm{\varepsilon}^{te}(\bm{X}^{tr}|J)&:= (\varepsilon^{te}(\bm{x}^{tr}_1|J), \ldots, \varepsilon^{te}(\bm{x}_n|J))^\top = (\bm{X}^{tr} - \bm{1}_n (\bm{\mu}^{te})^\top)(\bm{\beta} - \bm{\beta}^\star(J)). 
    \end{align}
    %
    
    \begin{lemma}
        \label{lem:lem5}
        Let $\bm{W}$ be the diagonal matrix with $w_t$'s as entries. 
        Under the same assumptions in Theorem \ref{thm:thm1}, we have
        \begin{align}
            &P\left(\bigcap_{k=1}^{K_n} \left\{\left|\frac{1}{n} \bm{\varepsilon}^{te}(\bm{X}^{tr}|\hat{J}_k)^\top \bm{W} \bm{\varepsilon}^{te}(\bm{X}^{tr}|\hat{J}_k) - E_n[\varepsilon^{te}(\bm{x}^{te}|\hat{J}_k)^2]\right| \right. \right. \\
            &\phantom{P\left(\bigcap_{k=1}^{K_n} \left\{\right. \right. } \left. \left. > \overline{B}d_n E_n[\varepsilon^{te}(\bm{x}^{te}|\hat{J}_k)^2]^{\frac{2\xi}{1+2\xi}}\right\}
            \right) = o (1). 
        \end{align}
    \end{lemma}

    {\sc proof}.
    By Lemma \ref{lem:lem4}, for any model $J \subset J_F$, 
    \begin{align}
        &\left| \frac{1}{n} \bm{\varepsilon}^{te}(\bm{X}^{tr}|J) ^\top \bm{W} \bm{\varepsilon}^{te}(\bm{X}^{tr}|J) - E_n[ \varepsilon^{te}(\bm{x}^{te}|J)^2 ] \right|  \\
        &\leq \left(\max\limits_{1\leq i \leq j \leq p_n} | \hat{\sigma}^{te}_{ij} - \sigma^{te}_{ij}| + \Vert \hat{\bm{\mu}}^{te} - \bm{\mu}^{te} \Vert_\infty^2 \right) \Vert \bm{\beta} - \bm{\beta}^\star(J) \Vert_1^2 \\
        &\leq C\left(\max\limits_{1\leq i \leq j \leq p_n} | \hat{\sigma}^{te}_{ij} - \sigma^{te}_{ij}| + \Vert \hat{\bm{\mu}}^{te} - \bm{\mu}^{te} \Vert_\infty^2\right) E_n[\varepsilon^{te}(\bm{x}^{te}|J)^2]^{\frac{2\xi}{1+2\xi}}. 
    \end{align}
    %


    Then the desired result follows immediately from \eqref{ing0105} and \eqref{imr250430}.    

    \begin{lemma}
        \label{lem:lem6}
        Under the same assumptions in Theorem \ref{thm:thm1}, we have
        \begin{align}
            P\left(
                \bigcap_{k=1}^{K_n}
                \left\{\left| \frac{1}{n}\bm{\varepsilon}^{te}(\bm{X}^{tr}|\hat{J}_k)^\top \bm{W} \bm{\varepsilon}^{tr}\right|
                > \overline{B} d_n E_n[\varepsilon^{te}(\bm{x}^{te}|\hat{J}_k)^2]^{\frac{\xi}{1+2\xi}}\right\}
            \right)
                &= o (1),
        \end{align}
        where $\bm{\varepsilon}^{tr} = (\varepsilon^{tr}_1, \ldots, \varepsilon^{tr}_n)^\top$. 
    \end{lemma}

    {\sc proof}.
    By Lemma \ref{lem:lem4}, for all $J \subset J_F$, 
    \begin{align}
        \left| \frac{1}{n} \bm{\varepsilon}^{te}(\bm{X}^{tr}|J) ^\top \bm{W} \bm{\varepsilon}^{tr} \right| 
        &\leq (1 + \Vert \bm{\mu}^{te} \Vert_\infty) \max\limits_{0\leq i \leq p_n} \left| \frac{1}{n} \sum\limits_{t=1}^n w_t x^{tr}_{ti} \varepsilon^{tr}_t \right| \Vert\bm{\beta} - \bm{\beta}^\star(J)\Vert_1\\
        &\leq C(1 + \Vert \bm{\mu}^{te} \Vert_\infty) \max\limits_{0\leq i \leq p_n} \left| \frac{1}{n} \sum\limits_{t=1}^n w_t x^{tr}_{ti} \varepsilon^{tr}_t \right| E_n[\varepsilon^{te}(\bm{x}^{te}|\hat{J}_k)^2]^{\frac{\xi}{1+2\xi}}. 
    \end{align}


    Hence, (C\ref{cond:cond2}), \eqref{ing0105}, and \eqref{imr250430-2} indicates the proof is completed.

    \begin{lemma}
        \label{lem:lem7}
        Under the same assumptions in Theorem \ref{thm:thm1}, we have
        \begin{align}
            P\left(
                \bigcap_{k=1}^{K_n}\left\{A(\hat{J}_k)
                > \overline{B} k d_n^2 E_n[\varepsilon^{te}(\bm{x}^{te}|\hat{J}_k)^2]^{\frac{2\xi}{1+2\xi}}
                \right\}
            \right)
                = o(1), 
        \end{align}
        where $A(J) = \bm{\varepsilon}^{te}(\bm{X}^{tr}|J)^\top \bm{W}^{1/2}  \bm{P}(J) \bm{W}^{1/2} \bm{\varepsilon}^{te}(\bm{X}^{tr}|J)/n$.
    \end{lemma}

    {\sc proof}.
    For all $J \subset J_F$ with $1 \leq \sharp{J} \leq K_n$, it follows from $\bm{P}_{\bm{W}}^{\perp}\bm{W}^{1/2} \bm{1}_n = \bm{0}_n$ and $(\bm{\Sigma}^{te}_{\cdot, J})^\top (\bm{\beta} - \bm{\beta}^\star(J)) = \bm{0}_{p_n}$ that 
    \begin{align}
        &\left\Vert \frac{1}{n} (\bm{X}^{te}_{\cdot,J})^\top \bm{W}^{1/2} \bm{P}_{\bm{W}}^{\perp}\bm{W}^{1/2} \bm{\varepsilon}^{te}(\bm{X}^{tr}|J) \right\Vert_2^2\\
        &= \Vert (\hat{\bm{\Sigma}}^{te}_{\cdot,J} - \bm{\Sigma}^{te}_{\cdot, J})^\top (\bm{\beta} - \bm{\beta}^\star(J)) \Vert_2^2\\
        &\leq \sharp{J}  \max\limits_{1\leq i \leq j \leq p_n} | \hat{\sigma}^{te}_{ij} - \sigma^{te}_{ij} |^2 \Vert\bm{\beta} - \bm{\beta}^\star(J)\Vert_1^2. 
        \label{imr250506-3}
    \end{align}
    Thus, we have 
    \begin{align}
        A(J) &\leq  \Vert(\hat{\bm{\Sigma}}^{te}_{J, J})^{-1}\Vert_2\left\Vert \frac{1}{n} (\bm{X}^{te}_{\cdot,J})^\top \bm{W}^{1/2} \bm{P}_{\bm{W}}^{\perp}\bm{W}^{1/2} \bm{\varepsilon}^{te}(\bm{X}^{tr}|J) \right\Vert_2^2\\
        &\leq \sharp{J} \Vert(\hat{\bm{\Sigma}}^{te}_{J, J})^{-1} \Vert_2 \max\limits_{1\leq i \leq j \leq p_n} | \hat{\sigma}^{te}_{ij} - \sigma^{te}_{ij} |^2 \Vert\bm{\beta} - \bm{\beta}^\star(J)\Vert_1^2.  
    \end{align}
    Then, the desired conclusion follows from Lemma \ref{lem:lem3}, Lemma \ref{lem:lem4}, and \eqref{imr250430}.
    
    \begin{lemma}
        \label{lem:lem8}
        Under the same assumptions in Theorem \ref{thm:thm1}, we have
            \begin{align}
                P\left( \bigcap_{k = 1}^{K_n}\left\{ B(\hat{J}_k) > \overline{B}k d_n^2 \right\}\right)
                    = o(1), 
            \end{align}
            where $B(J) = (\bm{\varepsilon}^{tr})^\top \bm{W}^{1/2} \bm{P}(J)\bm{W}^{1/2}\bm{\varepsilon}^{tr}/n$.
    \end{lemma}

    {\sc proof}.
    For all $J \subset J_F$, 
    \begin{align}
        & \left\Vert \frac{1}{n}(\bm{X}^{tr}_{\cdot,J})^\top\bm{W}^{1/2}\bm{P}_{\bm{W}}^\perp \bm{W}^{1/2} \bm{\varepsilon}^{tr} \right\Vert_2^2\\
        &\leq 2 \sharp{J} \left( \max_{1 \leq i \leq p_n}\left| \frac{1}{n} \sum_{t = 1}^n w_tx^{tr}_{ti}\varepsilon^{tr}_t \right|^2 + \Vert\hat{\bm{\mu}}^{te}\Vert_{\infty}^2 \left|\frac{1}{n}\sum_{t = 1}^n w_t \varepsilon^{tr}_t\right|^2 \right) \\
        &\leq 2 \sharp{J} (1 + \Vert\hat{\bm{\mu}}^{te}\Vert_{\infty}^2) \max\limits_{0 \leq i \leq p_n} \left| \frac{1}{n} \sum\limits_{t=1}^n w_t x^{tr}_{ti} \varepsilon^{tr}_t \right|^2. 
        \label{imr250506-4}
    \end{align}


    Thus, for $J \subset J_F$ with $1 \leq \sharp{J} \leq K_n$, 
    \begin{align}
        B(J) &\leq \max\limits_{1\leq \sharp{J} \leq K_n} \left\Vert ( \hat{\bm{\Sigma}}_{J,J}^{te})^{-1} \right\Vert_2 
        \left\Vert \frac{1}{n}(\bm{X}^{tr}_{\cdot,J})^\top\bm{W}^{1/2}\bm{P}_{\bm{W}}^\perp \bm{W}^{1/2} \bm{\varepsilon}^{tr} \right\Vert_2^2  \\
        &\leq 2 \sharp{J}  (1 + \Vert\hat{\bm{\mu}}^{te} \Vert_{\infty}^2) \max\limits_{1\leq \sharp{J} \leq K_n} \left\Vert ( \hat{\bm{\Sigma}}_{J,J}^{te})^{-1} \right\Vert_2 \max\limits_{0 \leq i \leq p_n} \left| \frac{1}{n} \sum\limits_{t=1}^n w_t x^{tr}_{ti} \varepsilon^{tr}_t \right|^2. 
    \end{align}
    Then, the desired result is readily seen by Lemma \ref{lem:lem3}, \eqref{ing0105}, and \eqref{imr250430-2}.

    \begin{lemma}
        \label{lem:lem9}
        Under the same assumptions in Theorem \ref{thm:thm1}, we have
        \begin{align}
            P\left( \bigcap_{k = 1}^{K_n}\left\{ \left| \frac{1}{n} \bm{1}_n^\top \bm{W} \bm{\varepsilon}^{te}(\bm{X}^{tr}|\hat{J}_k) \right| > \overline{B}d_n E_n[\varepsilon^{te}(\bm{x}^{te}|\hat{J}_k)^2]^{\frac{\xi}{1+2\xi}}\right\} \right)
                &= o(1). 
        \end{align}
    \end{lemma}

    {\sc proof}.
    For all $J \subset J_F$, 
    \begin{align}
        \left| \frac{1}{n} \bm{1}_n^\top \bm{W} \bm{\varepsilon}^{te}(\bm{X}^{tr}|J)  \right| &= |(\hat{\bm{\mu}}^{te} - \bm{\mu}^{te})^\top (\bm{\beta} - \bm{\beta}^\star(J))|  \\
        &\leq \Vert \hat{\bm{\mu}}^{te} - \bm{\mu}^{te} \Vert_\infty \Vert\bm{\beta} - \bm{\beta}^\star(J)\Vert_1. 
        \label{imr250506-7}
    \end{align}
    Thus, from Lemma \ref{lem:lem4} and \eqref{ing0105}, the desired result follows.


   \begin{lemma}
        \label{lem:lem10}
        Under the same assumptions in Theorem \ref{thm:thm1}, we have
        \begin{align}
            &P\left(\bigcap_{k = 1}^{K_n}\left\{ \Vert \hat{\bm{\beta}}(\hat{J}_k) - \bm{\beta}(\hat{J}_k) \Vert_2^2 > \overline{B}k d_n^2 \right\}\right)
            = o(1), \label{eq:eq20} \\
            &P\left(\bigcap_{k = 1}^{K_n}\left\{(\hat{\bm{\beta}}(\hat{J}_k) - \bm{\beta}(\hat{J}_k))^\top \bm{\Sigma}^{te}_{\hat{J}_k, \hat{J}_k} (\hat{\bm{\beta}}(\hat{J}_k) - \bm{\beta}(\hat{J}_k)) > \overline{B} k d_n^2\right\}\right)
            = o(1). \label{imr250505-2}
        \end{align}
    \end{lemma}
    
    {\sc proof}. 
    For all $J \subset J_F$ with $1 \leq \sharp{J} \leq K_n$, 
    \begin{align}
        \Vert \hat{\bm{\beta}}(J) - \bm{\beta}(J) \Vert_2^2 
        &= \left\Vert (\hat{\bm{\Sigma}}^{te}_{J,J})^{-1} \left( \frac{1}{n}(\bm{X}^{tr}_{\cdot,J})^\top\bm{W}^{1/2}\bm{P}_{\bm{W}}^\perp\bm{W}^{1/2}(\bm{\varepsilon}^{te}(\bm{X}^{tr}|J) + \bm{\varepsilon}^{tr})\right) \right\Vert_2^2  \\
        &\leq 2 \max_{1 \leq \sharp{J} \leq K_n}\Vert (\hat{\bm{\Sigma}}^{te}_{J,J})^{-1} \Vert_2^2 \left( \left\Vert \frac{1}{n}(\bm{X}^{tr}_{\cdot,J})^\top\bm{W}^{1/2}\bm{P}_{\bm{W}}^\perp\bm{W}^{1/2} \bm{\varepsilon}^{te}(\bm{X}^{tr}|J)
        \right\Vert_2^2 \right. \\
        &\phantom{= + 2 \max_{1 \leq \sharp{J} \leq K_n}\Vert (\hat{\bm{\Sigma}}^{te}_{J,J})^{-1} \Vert_2^2} 
        \left.  + \left\Vert \frac{1}{n}(\bm{X}^{tr}_{\cdot,J})^\top\bm{W}^{1/2}\bm{P}_{\bm{W}}^\perp\bm{W}^{1/2}\bm{\varepsilon}^{tr} \right\Vert_2^2\right). 
    \end{align}
    Note that (C\ref{cond:cond6}), \eqref{eq:eq52}, and \eqref{imr250506-5} indicate that 
    \begin{align}
        \max_{1 \leq \sharp{J} \leq K_n} \Vert\bm{\beta} - \bm{\beta}^\star(J)\Vert_1 \leq C. 
        \label{imr250506-8}
    \end{align}
    Combined this with \eqref{ing0105}--\eqref{imr250430}, \eqref{imr250506-3}, and \eqref{imr250506-4}, we obtain \eqref{eq:eq20}.

    On the other hand, 
    \begin{align}
        &(\hat{\bm{\beta}}(J) - \bm{\beta}(J))^\top \bm{\Sigma}^{te}_{J, J} (\hat{\bm{\beta}}(J) - \bm{\beta}(J)) \\
        &= (\hat{\bm{\beta}}(J) - \bm{\beta}(J))^\top \hat{\bm{\Sigma}}^{te}_{J, J} (\hat{\bm{\beta}}(J) - \bm{\beta}(J))\\
        &\phantom{=} + (\hat{\bm{\beta}}(J) - \bm{\beta}(J))^\top (\bm{\Sigma}^{te}_{J, J} - \hat{\bm{\Sigma}}^{te}_{J, J}) (\hat{\bm{\beta}}(J) - \bm{\beta}(J))\\
        &\leq \Vert (\hat{\bm{\Sigma}}^{te}_{J,J})^{-1} \Vert_2 \left\Vert \frac{1}{n}(\bm{X}^{tr}_{\cdot,J})^\top\bm{W}^{1/2}\bm{P}_{\bm{W}}^\perp\bm{W}^{1/2}(\bm{\varepsilon}^{te}(\bm{X}^{tr}|J) + \bm{\varepsilon}^{tr}) \right\Vert_2^2  \\
        &\phantom{\leq} + \sharp{J} \max_{1 \leq i \leq j \leq p_n}|\hat{\sigma}^{te}_{ij} - \sigma^{te}_{ij} | \Vert \hat{\bm{\beta}}(J) - \bm{\beta}(J) \Vert_2^2. 
    \end{align}
    Because $\sharp{J} \leq K_n = M_k d_n^{-1}$, \eqref{eq:eq20} and similar arguments to derive \eqref{eq:eq20} yield \eqref{imr250505-2}.

    \begin{lemma}
        \label{lem:lem11}
        Under the same assumptions in Theorem \ref{thm:thm1}, we have
        \begin{align}
            P\left( \bigcap_{k = 1}^{K_n} \left\{ ( \hat{\bm{\beta}}(\hat{J}_k)^\top ( \bm{\mu}^{te}_{\hat{J}_k} - \hat{\bm{\mu}}^{te}_{\hat{J}_k} ) + \hat{\mu}^{te}_y - \mu^{te}_y  )^2 > \overline{B}d_n^2 \right\}\right)
                &= o (1). 
        \end{align}
    \end{lemma}

    {\sc proof}.
    For a model $J \subset J_F$ with $1 \leq \sharp{J} \leq K_n$, 
    \begin{align}
        &\hat{\bm{\beta}}(J)^\top ( \bm{\mu}^{te}_J - \hat{\bm{\mu}}^{te}_J ) + \hat{\mu}^{te}_y - \mu^{te}_y \\
        &= (\bm{\beta}(J) - \hat{\bm{\beta}}(J) )^\top (\hat{\bm{\mu}}^{te}_J - \bm{\mu}^{te}_J )
        + \frac{1}{n} \bm{1}_n^\top  \bm{W} \bm{\varepsilon}^{te}(\bm{X}^{tr}|J)  + \frac{1}{n} \sum\limits_{t=1}^n w_t \varepsilon^{tr}_t. 
    \end{align}

    
    Then, 
    \begin{align}
        & ( \hat{\bm{\beta}}(J)^\top ( \bm{\mu}^{te}_J - \hat{\bm{\mu}}^{te}_J ) + \hat{\mu}^{te}_y - \mu^{te}_y )^2  \\
        &\leq 3
            \sharp{J} \Vert \bm{\beta}(J) - \hat{\bm{\beta}}(J) \Vert_2^2 \Vert \hat{\bm{\mu}}^{te} - \bm{\mu}^{te}\Vert_\infty^2 
            +3\left| \frac{1}{n}\bm{1}_n^\top  \bm{W} \bm{\varepsilon}^{te}(\bm{X}^{tr}|J)  \right|^2
             \\
        &\phantom{\leq }
            + 3 \max\limits_{0\leq i \leq p_n} \left| \frac{1}{n} \sum\limits_{t=1}^n w_t x^{tr}_{ti} \varepsilon^{tr}_t \right|^2. 
    \end{align}
    Because $\sharp{J} \leq K_n = M_Kd_n^{-1}$, by \eqref{ing0105}, \eqref{imr250506-7}, \eqref{eq:eq20}, and \eqref{imr250506-8}, the proof is completed. 
    
    \begin{remark}
        \label{rmk:rmk3}
        For Lemma \ref{lem:lem5} -- Lemma \ref{lem:lem11}, the considered model $\hat{J}_k$ can be actually replaced by any candidate model with cardinality $k$. 
    \end{remark}

    \begin{lemma}
        \label{lem:lem12} 
        Assume (C\ref{cond:cond7}) and \eqref{eq:eq61} hold. Then for all $J \subset J_F$, 
        \begin{align}
            \Vert \bm{\beta} - \bm{\beta}^{tr\star}(J) \Vert_1 \leq C \Vert \bm{\beta}_{J^c} \Vert_1, 
        \end{align}
        where let $\bm{\beta}^{tr\star}(J)\in\mathbb{R}^{p_n}$ be the augmented vector of $\bm{\beta}^{tr}(J) \in \mathbb{R}^{\sharp{J}}$, which is the coefficients of the best linear predictor for a specific model $J \subset J_F$ in the training population. 
    \end{lemma}

    {\sc proof}. 
    The assertion follows from a similar argument of \eqref{imr250506-5}. 


\subsection{Proof of Theorem \ref{thm:thm2} and \ref{thm:thm3}}
    \label{sec:sec6}
{\sc proof of Theorem \ref{thm:thm2}}. 
    We first express \eqref{eq:eq1} as
    \begin{align}
        \begin{split}
            (\hat{\sigma}^{te}(J))^2 &= \frac{1}{n} (\bm{\varepsilon}^{te}(\bm{X}^{tr}|J) + \bm{\varepsilon}^{tr})^\top \bm{W} (\bm{\varepsilon}^{te}(\bm{X}^{tr}|J) + \bm{\varepsilon}^{tr})  \\
            &\phantom{=}~~~~~-\frac{1}{n} (\bm{\varepsilon}^{te}(\bm{X}^{tr}|J) + \bm{\varepsilon}^{tr})^\top \bm{W}^{1/2} \bm{P}_{\bm{W}}\bm{W}^{1/2} (\bm{\varepsilon}^{te}(\bm{X}^{tr}|J) + \bm{\varepsilon}^{tr})  \\
            &\phantom{=}~~~~~-\frac{1}{n} (\bm{\varepsilon}^{te}(\bm{X}^{tr}|J) + \bm{\varepsilon}^{tr})^\top \bm{W}^{1/2} \bm{P}(J)\bm{W}^{1/2} (\bm{\varepsilon}^{te}(\bm{X}^{tr}|J) + \bm{\varepsilon}^{tr}). 
        \end{split}
        \label{eq:eq3}
    \end{align}
    %
    For simplicity, we use $\hat{\sigma}^2(J)$ instead of $(\hat{\sigma}^{te}(J))^2$ in the following. 
    In addition, hereafter, we consider $n$ is large enough in order to guarantee the events we consider occur with (any) high probability. 


    Let $m_n = \lfloor d_n^{-\frac{1}{1+\xi}} \rfloor - 1$. 
    Because $d_n \rightarrow 0$ under (C\ref{cond:cond1}), hereafter, we suppose that $n$ is large enough satisfying $1 \leq m_n \leq K_n = M_K d_n^{-1}$. 
    Note that $k^{-(1+2\xi)}\geq d_n^{\frac{1+2\xi}{1+\xi}}$ if $k \leq m_n + 1$. 
    It follows from \eqref{eq:eq1033} that for any small $\varepsilon > 0$, there exist $C_\varepsilon , N_{\varepsilon} > 0$ such that for all $n \geq N_{\varepsilon}$, 
    \begin{align}
        \mathcal{E}_{n, \varepsilon} := \bigcap_{k = 1}^{K_n}\left\{E_n[\varepsilon^{te}(\bm{x}^{te}|\hat{J}_k)^2] \leq \frac{C_\varepsilon }{2}\left( k^{-(1+2\xi)} + d_n^{\frac{1+2\xi}{1+\xi}} \right)\right\} 
        \label{imr250505-1}
    \end{align}
    satisfies 
    \begin{align}
    \label{ing250050}
        P(\mathcal{E}_{n, \varepsilon}) \geq 1 - \varepsilon. 
    \end{align} 
    It is clear that $\mathcal{E}_{n, \varepsilon}$ implies
    \begin{align}
        E_n[\varepsilon^{te}(\bm{x}^{te}|\hat{J}_{m_n})^2] \leq C_\varepsilon  m_n^{-(1+2\xi)}. 
        \label{eq:eq4}
    \end{align}

    
    %
    Define
    \begin{align}
        \tilde{k} := \min\left\{ k \in \{1, \ldots, K_n\} \mid E_n[\varepsilon^{te}(\bm{x}^{te}|\hat{J}_k)^2] \leq G_\varepsilon  m_n^{-(1+2\xi)} \right\}, \label{eq:eq5}
    \end{align}
    where $G_\varepsilon >C_\varepsilon $ and $\min \emptyset = K_n$. 
    Note that
    \eqref{eq:eq4} yields
    $1 \leq \tilde{k} \leq m_n$ on $\mathcal{E}_{n, \varepsilon}$. 
    The proof can be  divided into three steps:
    \begin{align}
        &{\rm I}.~P(\hat{k}\leq \tilde{k}-1) = o(1). \label{eq:eq6} \\
        &{\rm II}.~P(\hat{k} > V_\varepsilon  m_n) = o(1),~\text{for a large constant satisfying $V_\varepsilon \geq 2$}. \label{eq:eq7} \\
        &{\rm III}.~E_n[ (y^{te}(\bm{x}^{te}) - \hat{y}^{te}(\bm{x}^{te}|\hat{J}_{\hat{k}}))^2 ] I_{\{\tilde{k}\leq \hat{k}\leq V_\varepsilon  m_n \}} = O_p \left( d_n^{\frac{1+2\xi}{1+\xi}} \right). \label{eq:eq8}
    \end{align}

    {\sc step I: proof of \eqref{eq:eq6}}. \\
    Since the $\varepsilon$ in \eqref{ing250050}
    can be made arbitrarily small, we may assume,
    without loss of generality, that
    $\mathcal{E}_{n, \varepsilon}$ holds. 
    In addition, we focus on the non-trivial case
    where $2 \leq \tilde{k} \leq m_n$. 
    Then, 
    \begin{align}
    \begin{split}
        &\{ \hat{k} \leq \tilde{k}-1 \}
            \subseteq \bigcup\limits_{k=1}^{\tilde{k}-1} \left\{ \text{HDIWIC}(\hat{J}_k) \leq \text{HDIWIC}(\hat{J}_{m_n}) \right\}  \\
            &\subseteq \bigcup\limits_{k=1}^{\tilde{k}-1} \left\{ \left( 1 + s_a (k+1) d_n^2 \right) \hat{\sigma}^2(\hat{J}_k) \leq \left( 1 + s_a (m_n+1) d_n^2 \right) \hat{\sigma}^2(\hat{J}_{m_n}) \right\}  \\
            &\subseteq \bigcup\limits_{k=1}^{\tilde{k}-1} \left\{ \hat{\sigma}^2(\hat{J}_k) - \hat{\sigma}^2(\hat{J}_{m_n}) \leq s_a (m_n+1) d_n^2 \hat{\sigma}^2(\hat{J}_{m_n}) \right\}.  \label{eq:eq10}
    \end{split}
    \end{align}
    From \eqref{eq:eq3}, 
    \begin{align}
        \begin{split}
            \hat{\sigma}^2(\hat{J}_k) - \hat{\sigma}^2(\hat{J}_{m_n}) 
            &\geq \frac{1}{n} \bm{\varepsilon}^{te}(\bm{X}^{tr}|\hat{J}_k)^\top \bm{W} \bm{\varepsilon}^{te}(\bm{X}^{tr}|\hat{J}_k)\\
            &\phantom{\geq}+ \frac{2}{n} \bm{\varepsilon}^{te}(\bm{X}^{tr}|\hat{J}_k)^\top \bm{W} \bm{\varepsilon}^{tr}  \\
            &\phantom{\geq}- \frac{1}{n}\bm{\varepsilon}^{te}(\bm{X}^{tr}|\hat{J}_k)^\top \bm{W}^{1/2} \bm{P}_{\bm{W}} \bm{W}^{1/2} \bm{\varepsilon}^{te}(\bm{X}^{tr}|\hat{J}_k)   \\
            &\phantom{\geq}- \frac{2}{n} \bm{\varepsilon}^{te}(\bm{X}^{tr}|\hat{J}_k)^\top \bm{W}^{1/2} \bm{P}_{\bm{W}} \bm{W}^{1/2} \bm{\varepsilon}^{tr} \\
            &\phantom{\geq}- 2(A(\hat{J}_k) + B(\hat{J}_k))  \\
            &\phantom{\geq}-\frac{1}{n} \bm{\varepsilon}^{te}(\bm{X}^{tr}|\hat{J}_{m_n})^\top \bm{W} \bm{\varepsilon}^{te}(\bm{X}^{tr}|\hat{J}_{m_n}) \\
            &\phantom{\geq}- \frac{2}{n} \bm{\varepsilon}^{te}(\bm{X}^{tr}|\hat{J}_{m_n})^\top \bm{W} \bm{\varepsilon}^{tr} \\
            &\phantom{\geq}+ \frac{2}{n} \bm{\varepsilon}^{te}(\bm{X}^{tr}|\hat{J}_{m_n})^\top \bm{W}^{1/2} \bm{P}_{\bm{W}} \bm{W}^{1/2} \bm{\varepsilon}^{tr}, 
        \end{split}
        \label{eq:eq11}
    \end{align}
    where $A(J)$ and $B(J)$ are defined in Lemma \ref{lem:lem7} and Lemma \ref{lem:lem8}, respectively.

    We evaluate the components on the right-hand side of \eqref{eq:eq11} as follows. \\
    By Lemma \ref{lem:lem5} and \eqref{eq:eq5}, 
    it holds that, with high probability, for all $k=1, \ldots, \tilde{k}-1$, 
    \begin{align}
        &\frac{1}{n} \bm{\varepsilon}^{te}(\bm{X}^{tr}|\hat{J}_k)^\top \bm{W} \bm{\varepsilon}^{te}(\bm{X}^{tr}|\hat{J}_k)  \\
        &\geq E_n[\varepsilon^{te}(\bm{x}^{te}|\hat{J}_k)^2] - \left| \frac{1}{n} \bm{\varepsilon}^{te}(\bm{X}^{tr}|\hat{J}_k)^\top \bm{W} \bm{\varepsilon}^{te}(\bm{X}^{tr}|\hat{J}_k) - E_n [\bm{\varepsilon}^{te}(\bm{X}^{tr}|\hat{J}_k)^2] \right|  \\
        &\geq E_n[\varepsilon^{te}(\bm{x}^{te}|\hat{J}_k)^2] \left( 1 - \overline{B} G_\varepsilon^{-\frac{1}{1+2\xi}} d_n^{\frac{\xi}{1 + \xi}} \right).
        \label{eq:eq23}
    \end{align}
    %
    Lemma \ref{lem:lem6} and \eqref{eq:eq5}
    ensure that, with high probability, for all $k = 1, \ldots, \tilde{k}-1$, 
    \begin{align}
        \frac{2}{n} \bm{\varepsilon}^{te}(\bm{X}^{tr}|\hat{J}_k)^\top \bm{W} \bm{\varepsilon}^{tr} 
        \geq - 2\overline{B} G_\varepsilon^{-\frac{1+\xi}{1+2\xi}} E_n[\varepsilon^{te}(\bm{x}^{te}|\hat{J}_k)^2]. 
        \label{eq:eq24}
    \end{align}
    %
    By Lemma \ref{lem:lem9} and \eqref{eq:eq5}, it holds that, with high probability, for all $k = 1, \ldots, \tilde{k}-1$, 
    \begin{align}
        -\frac{1}{n}\bm{\varepsilon}^{te}(\bm{X}^{tr}|\hat{J}_k)^\top \bm{W}^{1/2} \bm{P}_{\bm{W}} \bm{W}^{1/2} \bm{\varepsilon}^{te}(\bm{X}^{tr}|\hat{J}_k)
        & = - \left| \frac{1}{n} \bm{1}_n^\top \bm{W} \bm{\varepsilon}^{te}(\bm{X}^{tr}|\hat{J}_k)\right|^2 \\
        &\geq - \overline{B} G_\varepsilon^{-\frac{1}{1+2\xi}} d_n^{\frac{1+2\xi}{1+\xi}} E_n[\varepsilon^{te}(\bm{x}^{te}|\hat{J}_k)^2]. 
        \label{eq:eq25} 
    \end{align}
    %
    By Lemma \ref{lem:lem9}, \eqref{ing0105}, and \eqref{eq:eq5}, it holds that, with high probability, for all $k = 1, \ldots, \tilde{k}-1$, 
    \begin{align}
        -\frac{2}{n} \bm{\varepsilon}^{te}(\bm{X}^{tr}|\hat{J}_k)^\top \bm{W}^{1/2} \bm{P}_{\bm{W}} \bm{W}^{1/2} \bm{\varepsilon}^{tr}
        &\geq - 2 \left| \frac{1}{n}\bm{1}_n^\top \bm{W} \bm{\varepsilon}^{te}(\bm{X}^{tr}|\hat{J}_k) \right| \left|\frac{1}{n} \bm{1}_n^\top \bm{W} \bm{\varepsilon}^{tr}\right|\\
        &\geq - 2\overline{B}^2 G_\varepsilon^{-\frac{1+\xi}{1+2\xi}} d_n E_n[\varepsilon^{te}(\bm{x}^{te}|\hat{J}_k)^2]. 
        \label{eq:eq26} 
    \end{align}
    %
    By Lemma \ref{lem:lem7}, Lemma \ref{lem:lem8}, and \eqref{eq:eq5}, it holds that, with high probability, for all $k = 1, \ldots, \tilde{k} - 1$, 
    \begin{align}
        - 2 (A(\hat{J}_k) + B(\hat{J}_k) )
        \geq - 2\overline{B}\left(G_\varepsilon^{-\frac{1}{1 + 2\xi}} d_n^{\frac{2\xi}{1 + \xi}} + G_\varepsilon^{-1}\right) E_n[\varepsilon^{te}(\bm{x}^{te}|\hat{J}_k)^2]. 
        \label{eq:eq27}
    \end{align}
    %
    By Lemma \ref{lem:lem5}, \eqref{eq:eq4}, and \eqref{eq:eq5}, it holds that, with high probability, for all $k = 1, \ldots, \tilde{k} - 1$, 
    \begin{align}
        &-\frac{1}{n} \bm{\varepsilon}^{te}(\bm{X}^{tr}|\hat{J}_{m_n})^\top \bm{W} \bm{\varepsilon}^{te}(\bm{X}^{tr}|\hat{J}_{m_n}) \\
        &\geq -E_n[\varepsilon^{te}(\bm{x}^{te}|\hat{J}_{m_n})^2] -\left|\frac{1}{n} \bm{\varepsilon}^{te}(\bm{X}^{tr}|\hat{J}_k)^\top \bm{W} \bm{\varepsilon}^{te}(\bm{X}^{tr}|\hat{J}_{m_n}) - E_n[\varepsilon^{te}(\bm{x}^{te}|\hat{J}_{m_n})^2]\right| \\
        &\geq -G_\varepsilon^{-1} \left(C_\varepsilon + \overline{B} C_\varepsilon^{\frac{2\xi}{1+2\xi}} d_n^{\frac{\xi}{1 + \xi}} \right)E_n[\varepsilon^{te}(\bm{x}^{te}|\hat{J}_k)^2]. 
        \label{eq:eq28}
    \end{align}
   %
    By Lemmas \ref{lem:lem6}, \eqref{eq:eq4}, and \eqref{eq:eq5}, it holds with high probability that for all $k = 1, \ldots, \tilde{k} - 1$, 
    \begin{align}
        - \frac{2}{n} \bm{\varepsilon}^{te}(\bm{X}^{tr}|\hat{J}_{m_n})^\top \bm{W} \bm{\varepsilon}^{tr} &\geq - \overline{B} C_\varepsilon^{\frac{1+\xi}{1+2\xi}} G_\varepsilon^{-1} E_n[\varepsilon^{te}(\bm{x}^{te}|\hat{J}_k)^2]. 
        \label{eq:eq29}
    \end{align}
    %
Moreover, Lemma \ref{lem:lem9}, \eqref{ing0105}, \eqref{eq:eq4}, and \eqref{eq:eq5} imply that with high probability, for all $k = 1, \ldots, \tilde{k} - 1$, 
    \begin{align}
        \frac{2}{n} \bm{\varepsilon}^{te}(\bm{X}^{tr}|\hat{J}_{m_n})^\top \bm{W}^{1/2} \bm{P}_{\bm{W}} \bm{W}^{1/2} \bm{\varepsilon}^{tr}
        &\geq -2\left|\frac{1}{n}\bm{1}_n^\top \bm{W} \bm{\varepsilon}^{te}(\bm{X}^{tr}|\hat{J}_{m_n}) \right| \left| \frac{1}{n} \bm{1}_n^\top \bm{W} \bm{\varepsilon}^{tr} \right|\\
        &\geq -2\overline{B}^2 C_\varepsilon^{\frac{\xi}{1 + 2\xi}} G_\varepsilon^{-1} d_n E_n[\varepsilon^{te}(\bm{x}^{te}|\hat{J}_k)^2]. 
    \end{align}
    Combining these results, we obtain for sufficiently large $n$ and all $k = 1, \ldots, \tilde{k} - 1$, 
    \begin{align}
        \hat{\sigma}^2(\hat{J}_k) - \hat{\sigma}^2(\hat{J}_{m_n}) 
        \geq \frac{1}{2}E_n[ \varepsilon^{te}(\bm{x}^{te}|\hat{J}_k)^2 ] 
        \geq \frac{G_\varepsilon }{2} m_n^{-(1+2\xi)},
        \label{eq:eq30}
    \end{align}
with high probability.
In addition, using
    \begin{align}
        \frac{1}{n}\sum\limits_{t=1}^n w_t(\varepsilon^{tr}_t)^2 = \sigma^2 + o_p(1), \label{eq:eq9}
    \end{align}
and an argument similar to that used in the proof of 
\eqref{eq:eq30}, it can be shown that
$\hat{\sigma}^2(\hat{J}_{m_n}) \leq \overline{M}_1 + o_p(1)$ for some constant $\overline{M}_1>0$.
    Thus, with high probability, it holds that 
    \begin{align}
        s_a (m_n+1) d_n^2 \hat{\sigma}^2(\hat{J}_{m_n})
        \leq \overline{M}_1 s_a d_n^{\frac{2\xi+1}{\xi+1}}. 
        \label{eq:eq31}
    \end{align}
    Combining \eqref{eq:eq10}, \eqref{eq:eq30}, and \eqref{eq:eq31}, and letting $G_\varepsilon$
    be sufficiently large, we obtain
    \begin{align}
        P(\hat{k} \leq \tilde{k} - 1)  = o(1), 
        \label{eq:eq32}
    \end{align}
    which completes  the proof of \eqref{eq:eq6}. 


    {\sc step II: proof of  \eqref{eq:eq7}}. \\
    Under (C\ref{cond:cond1}), for all large $n$, 
    \begin{align}
        &\{ \hat{k} > V_\varepsilon  m_n \}  \\
        &\subseteq \bigcup\limits_{V_\varepsilon m_n < k \leq K_n} \left\{ \text{HDIWIC}(\hat{J}_k) \leq \text{HDIWIC}(\hat{J}_{m_n}) \right\}  \\
        &= \bigcup\limits_{V_\varepsilon m_n < k \leq K_n} \left\{ s_a (k - m_n) d_n^2 \hat{\sigma}^2(\hat{J}_k) \leq \left( 1 + s_a (m_n+1) d_n^2 \right) ( \hat{\sigma}^2(\hat{J}_{m_n}) - \hat{\sigma}^2(\hat{J}_k) ) \right\}  \\
        &\overset{}{\subseteq} \bigcup\limits_{V_\varepsilon m_n < k \leq K_n} \left\{ s_a (k - m_n) d_n^2 \hat{\sigma}^2(\hat{J}_k) \leq \frac{3}{2}( \hat{\sigma}^2(\hat{J}_{m_n}) - \hat{\sigma}^2(\hat{J}_k) ) \right\}.  \label{eq:eq33}
    \end{align}
    By \eqref{eq:eq3}, we have
    \begin{align}
    \begin{split}
        \hat{\sigma}^2(\hat{J}_{m_n}) - \hat{\sigma}^2(\hat{J}_k)
            &\leq \frac{1}{n} \bm{\varepsilon}^{te}(\bm{X}^{tr}|\hat{J}_{m_n})^\top \bm{W} \bm{\varepsilon}^{te}(\bm{X}^{tr}|\hat{J}_{m_n}) \\
            &\phantom{\leq} + \frac{2}{n} \bm{\varepsilon}^{te}(\bm{X}^{tr}|\hat{J}_{m_n})^\top \bm{W} \bm{\varepsilon}^{tr} \\
            & \phantom{\leq} - \frac{2}{n} \bm{\varepsilon}^{te}(\bm{X}^{tr}|\hat{J}_{m_n})^\top \bm{W}^{1/2} \bm{P}_{\bm{W}} \bm{W}^{1/2} \bm{\varepsilon}^{tr} \\
            &\phantom{\leq} -  \frac{2}{n} \bm{\varepsilon}^{te}(\bm{X}^{tr}|\hat{J}_k)^\top \bm{W} \bm{\varepsilon}^{tr}  \\
            &\phantom{\leq} + \frac{1}{n} \bm{\varepsilon}^{te}(\bm{X}^{tr}|\hat{J}_k)^\top \bm{W}^{1/2} \bm{P}_{\bm{W}} \bm{W}^{1/2} \bm{\varepsilon}^{te}(\bm{X}^{tr}|\hat{J}_k)  \\
            &\phantom{\leq} + \frac{2}{n} \bm{\varepsilon}^{te}(\bm{X}^{tr}|\hat{J}_k)^\top \bm{W}^{1/2} \bm{P}_{\bm{W}} \bm{W}^{1/2} \bm{\varepsilon}^{tr} \\
            &\phantom{\leq}+ 2 (A(\hat{J}_k) + B(\hat{J}_k)). 
    \end{split}
    \label{eq:eq34}
    \end{align}
    By Lemma \ref{lem:lem5} and \eqref{eq:eq4}, it holds that, with high probability, 
    \begin{align}
        &\frac{1}{n} \bm{\varepsilon}^{te}(\bm{X}^{tr}|\hat{J}_{m_n})^\top \bm{W} \bm{\varepsilon}^{te}(\bm{X}^{tr}|\hat{J}_{m_n})  \\
        &\leq E_n[\varepsilon^{te}(\bm{x}^{te}|\hat{J}_{m_n})^2] + \left| \frac{1}{n} \bm{\varepsilon}^{te}(\bm{X}^{tr}|\hat{J}_{m_n})^\top \bm{W} \bm{\varepsilon}^{te}(\bm{X}^{tr}|\hat{J}_{m_n}) - E_n[\varepsilon^{te}(\bm{x}^{te}|\hat{J}_{m_n})^2] \right|  \\
        &\leq C_\varepsilon m_n^{-(1 + 2\xi)}  + \overline{B} C_\varepsilon^{\frac{2\xi}{1+2\xi}} d_n m_n^{-2\xi}.
        \label{eq:eq35}
    \end{align}
    By Lemma \ref{lem:lem6} and \eqref{eq:eq4}, it holds that, with high probability, 
    \begin{align}
        \frac{2}{n} \bm{\varepsilon}^{te}(\bm{X}^{tr}|\hat{J}_{m_n})^\top \bm{W} \bm{\varepsilon}^{tr} \leq 2\overline{B}C_\varepsilon^{\frac{\xi}{1+2\xi}} d_n  m_n^{-\xi}. 
         \label{eq:eq36}
    \end{align}
    %
    By Lemma \ref{lem:lem6} and \eqref{imr250505-1}, it holds that, with high probability, for all $k > V_\varepsilon m_n$, 
    \begin{align}
        - \frac{2}{n} \bm{\varepsilon}^{te}(\bm{X}^{tr}|\hat{J}_k)^\top \bm{W} \bm{\varepsilon}^{te}
            &\leq 2\overline{B}C_\varepsilon^{\frac{\xi}{1+2\xi}} d_n m_n^{-\xi}. 
            \label{eq:eq59}
    \end{align}
    %
    By Lemma \ref{lem:lem9} and \eqref{imr250505-1}, it holds that, with high probability, for all $k > V_\varepsilon m_n$, 
    \begin{align}
        \frac{1}{n} \bm{\varepsilon}^{te}(\bm{X}^{tr}|\hat{J}_k)^\top \bm{W}^{1/2} \bm{P}_{\bm{W}} \bm{W}^{1/2} \bm{\varepsilon}^{te}(\bm{X}^{tr}|\hat{J}_k) &= \left| \frac{1}{n} \bm{1}_n^\top \bm{W} \bm{\varepsilon}^{te}(\bm{X}^{tr}|\hat{J}_k)\right|^2 \\
        &\leq \overline{B}^2 C_\varepsilon^{\frac{2\xi}{1+2\xi}} d_n^2  m_n^{-2\xi}. 
        \label{eq:eq37}
    \end{align}
    By Lemma \ref{lem:lem9}, \eqref{ing0105}, and \eqref{eq:eq4}, it holds that, with high probability, 
    \begin{align}
        -\frac{2}{n} \bm{\varepsilon}^{te}(\bm{X}^{tr}|\hat{J}_{m_n})^\top \bm{W}^{1/2} \bm{P}_{\bm{W}} \bm{W}^{1/2} \bm{\varepsilon}^{tr}
        &\leq 2 \left| \frac{1}{n}\bm{1}_n^\top \bm{W} \bm{\varepsilon}^{te}(\bm{X}^{tr}|\hat{J}_{m_n}) \right| \left|\frac{1}{n} \bm{1}_n^\top \bm{W} \bm{\varepsilon}^{tr}\right|\\
        &\leq 2\overline{B}^2 C_\varepsilon^{\frac{\xi}{1+2\xi}} d_n^2m_n^{-\xi}.
    \end{align}
    By Lemma \ref{lem:lem9}, \eqref{ing0105}, and \eqref{imr250505-1}, it holds that, with high probability, for  all $k > V_\varepsilon m_n$, 
    \begin{align}
        \frac{2}{n} \bm{\varepsilon}^{te}(\bm{X}^{tr}|\hat{J}_k)^\top \bm{W}^{1/2} \bm{P}_{\bm{W}} \bm{W}^{1/2} \bm{\varepsilon}^{tr}
        &\leq 2 \left| \frac{1}{n}\bm{1}_n^\top \bm{W} \bm{\varepsilon}^{te}(\bm{X}^{tr}|\hat{J}_k) \right| \left|\frac{1}{n} \bm{1}_n^\top \bm{W} \bm{\varepsilon}^{tr}\right|\\
        &\leq 2\overline{B}^2 C_\varepsilon^{\frac{\xi}{1+2\xi}} d_n^2m_n^{-\xi}.
        \label{eq:eq38}
    \end{align}
    By Lemmas \ref{lem:lem7}, \ref{lem:lem8}, and \eqref{imr250505-1}, it holds that, with high probability, for all $k > V_\varepsilon m_n$, 
    \begin{align}
        2 ( A(\hat{J}_k) + B(\hat{J}_k)) &\leq 2\overline{B}\left(1 +  C_\varepsilon^{\frac{1 + \xi}{1 + 2\xi}} m_n^{-(1 + \textcolor{magenta}{2}\xi)}\right) k d_n^2. 
        \label{eq:eq39}
    \end{align}
    %
    Since  $m_n \rightarrow \infty$ under (C\ref{cond:cond1}), 
    the above inequalities 
    imply that there exists $Q_\varepsilon > 0$, depending on $\overline{B}$ and $C_\varepsilon$,  such that, with high probability, for all $k > V_\varepsilon m_n$ and sufficiently large $n$, 
    \begin{align}
    \begin{split}
        \hat{\sigma}^2(\hat{J}_{m_n}) - \hat{\sigma}^2(\hat{J}_k) &\leq Q_\varepsilon m_n^{-(1 + 2\xi)} + 3\overline{B} k m_n^{-2(1 + \xi)}\\
        &\leq  (Q_\varepsilon V_\varepsilon^{-1} + 3\overline{B}) k m_n^{-2(1 + \xi)}. 
        \label{eq:eq40}
    \end{split}
    \end{align}
    
    By \eqref{eq:eq9} and an argument
    similar to that used in the proof of
    \eqref{eq:eq40}, there exists a constant $\overline{M}_2 > 0$ such that 
    \begin{align}
        P\left(\min_{k > V_\varepsilon m_n} \hat{\sigma}^2(\hat{J}_k) < \overline{M}_2\right) = o(1). 
        \label{eq:eq43}
    \end{align}
    Combine  \eqref{eq:eq33}, \eqref{eq:eq40}, and \eqref{eq:eq43}, we have
    \begin{align}
        &P(\hat{k} > V_\varepsilon m_n)  \\
         &\leq P\left(\bigcup\limits_{V_\varepsilon m_n < k \leq K_n}\left\{ \overline{M}_2 s_a (k-m_n) d_n^2  \leq \frac{3}{2} (Q_\varepsilon V_\varepsilon^{-1} + 3\overline{B}) k m_n^{-2(1 + \xi)} \right\}\right) + 2\varepsilon \\
        &\leq P\left(\bigcup\limits_{V_\varepsilon m_n < k \leq K_n}\left\{  \overline{M}_2 (1 - V_\varepsilon^{-1}) s_a  k d_n^2 \leq \frac{3}{2} (Q_\varepsilon V_\varepsilon^{-1} + 3\overline{B}) k m_n^{-2(1 + \xi)} \right\}\right) + 2\varepsilon \\
        &\leq P( \overline{M}_2s_a (1 + 2m_n^{-1})^{-2(1 + \xi)} \leq 3(Q_\varepsilon V_\varepsilon^{-1} + 3\overline{B})) + 2\varepsilon, 
    \end{align}
    where the last inequality follows from $V_\varepsilon \geq 2$ and $m_n \geq d_n^{-\frac{1}{1 + \xi}} - 2$. 
    Note that $\overline{B}$ is independent of $\varepsilon$ and hence we can pick $V_\varepsilon$ satisfying
    \begin{align}
        Q_\varepsilon V_\varepsilon^{-1} \leq \overline{B}.  
    \end{align}
    Thus, by choosing $s_a$ to satisfy $\overline{M}_2s_a > 12\overline{B}$, we obtain, for all sufficiently large $n$,
    \begin{align}
        P(\hat{k} > V_\varepsilon m_n) \leq 2\varepsilon. 
    \end{align}
    Because $\varepsilon > 0$ can be arbitrarily small, the proof of \eqref{eq:eq7} is complete. 
    
    {\sc step III: proof of \eqref{eq:eq8}}.\\
    Note first that
    \begin{align}
        &E_n[(y^{te}(\bm{x}^{te}) - \hat{y}^{te}(\bm{x}^{te}|\hat{J}_{\hat{k}}))^2] I_{\{\tilde{k}\leq \hat{k}\leq V_\varepsilon m_n \}}  \\
        &= E_n [\varepsilon^{te}(\bm{x}^{te}|\hat{J}_{\hat{k}})^2] I_{\{\tilde{k}\leq \hat{k}\leq V_\varepsilon m_n \}}  \\
        &\phantom{=} + E_n [ \{ (\hat{\alpha}(\hat{J}_{\hat{k}}) - \alpha(\hat{J}_{\hat{k}}) ) + ( \hat{\bm{\beta}}(\hat{J}_{\hat{k}}) - \bm{\beta}(\hat{J}_{\hat{k}}) )^\top \bm{x}^{te}_{\hat{J}_{\hat{k}}} \}^2 ] I_{\{\tilde{k}\leq \hat{k}\leq V_\varepsilon m_n \}}.
        \label{ing250040}
    \end{align}
    Because $E_n [\varepsilon^{te}(\bm{x}^{te}|\hat{J}_k)^2]$ decreases as $k$ increases, by \eqref{eq:eq5}, 
    \begin{align}
        E_n [\varepsilon^{te}(\bm{x}^{te}|\hat{J}_{\hat{k}})^2] I_{\{\tilde{k}\leq \hat{k}\leq V_\varepsilon m_n \}}
            &\leq E_n [\varepsilon^{te}(\bm{x}^{te}|\hat{J}_{\tilde{k}})^2] \leq G_\varepsilon  m_n^{-(1+2\xi)}. 
    \end{align}
    In addition, with high probability, 
    \begin{align}
        &E_n [ \{ (\hat{\alpha}(\hat{J}_{\hat{k}}) - \alpha(\hat{J}_{\hat{k}})) + (\hat{\bm{\beta}}(\hat{J}_{\hat{k}}) - \bm{\beta}(\hat{J}_{\hat{k}}))^\top \bm{x}^{te}_{\hat{J}_{\hat{k}}} \}^2 ] I_{\{\tilde{k}\leq \hat{k}\leq V_\varepsilon m_n \}}  \\
        &= E_n[ \{
                ( \hat{\bm{\beta}}^\top(\hat{J}_{\hat{k}}) ( \bm{\mu}^{te}_{\hat{J}_{\hat{k}}} - \hat{\bm{\mu}}^{te}_{\hat{J}_{\hat{k}}} ) + \hat{\mu}^{te}_y - \mu^{te}_y )  \\
        &\phantom{E_n \{} +
                ( \hat{\bm{\beta}}(\hat{J}_{\hat{k}}) - \bm{\beta}(\hat{J}_{\hat{k}}) )^\top ( \bm{x}^{te}_{\hat{J}_{\hat{k}}} - \bm{\mu}^{te}_{\hat{J}_{\hat{k}}} )
            \}^2 I_{\{\tilde{k}\leq \hat{k}\leq V_\varepsilon m_n \}}  \\
        &\leq 2 \{ \hat{\bm{\beta}}^\top(\hat{J}_{\hat{k}}) ( \bm{\mu}^{te}_{\hat{J}_{\hat{k}}} - \hat{\bm{\mu}}^{te}_{\hat{J}_{\hat{k}}} ) + \hat{\mu}^{te}_y - \mu^{te}_y \}^2 I_{\{\tilde{k}\leq \hat{k}\leq V_\varepsilon m_n \}}  \\
        &\phantom{\leq}~~~~~+ 2 E_n[ \{ ( \hat{\bm{\beta}}(\hat{J}_{\hat{k}}) - \bm{\beta}(\hat{J}_{\hat{k}}) )^\top (\bm{x}^{te}_{\hat{J}_{\hat{k}}} - \bm{\mu}^{te}_{\hat{J}_{\hat{k}}}) \}^2 ] I_{\{\tilde{k}\leq \hat{k}\leq V_\varepsilon m_n \}}  \\
        &\leq 2\overline{B}(1 + \hat{k})d_n^2I_{\{\tilde{k}\leq \hat{k}\leq V_\varepsilon m_n \}} , 
        \label{eq:eq45}
    \end{align}
    where the last inequality is ensured by Lemmas \ref{lem:lem10} and  \ref{lem:lem11}. 
    Hence, with high probability, we have
    \begin{align}
        &E_n [ \{ (\hat{\alpha}(\hat{J}_{\hat{k}}) - \alpha(\hat{J}_{\hat{k}})) + (\hat{\bm{\beta}}(\hat{J}_{\hat{k}}) - \bm{\beta}(\hat{J}_{\hat{k}}))^\top \bm{x}^{te}_{\hat{J}_{\hat{k}}} \}^2 ] I_{\{\tilde{k}\leq \hat{k} \leq V_\varepsilon m_n \}}\\
        &\leq 4\overline{B}V_{\varepsilon}m_n^{-(1 + 2\xi)}. 
        \label{eq:eq46}
    \end{align}
    Combine the above inequalities, we have, with high probability,
    \begin{align}
        E_n[(y^{te}(\bm{x}^{te}) - \hat{y}^{te}(\bm{x}^{te}|\hat{J}_{\hat{k}}))^2] I_{\{\tilde{k}\leq \hat{k}\leq V_\varepsilon m_n \}} \leq (G_\varepsilon + 4\overline{B}V_{\varepsilon})m_n^{-(1 + 2\xi)}. 
    \end{align}
    This completes the proof of \eqref{eq:eq8}.
    \hfill $\blacksquare$

{\sc proof of Theorem \ref{thm:thm3}}.
    Replace $w_{t}$'s, $\hat{\sigma}^{te}(J)$, $\hat{y}^{te}(\bm{x}|J)$, $\hat{\alpha}(J)$, $\hat{\beta}(J)$, $\text{HDIWIC}$ and $\hat{k}$ by $\hat{w}_{t}$'s, $\check{\sigma}^{te}(J)$, $\check{y}^{te}(\bm{x}|J)$, $\check{\alpha}(J)$, $\check{\beta}(J)$, $\text{HDIWIC}_{s}$ and $\check{k}$. Then, by \eqref{eq:eq55}, \eqref{eq:eq133} and Lemma \ref{lem:lem1}, we can assert that
    \begin{align}
        &P\left( \max\limits_{0 \leq i \leq j \leq p_n} \left| \frac{1}{n} \sum\limits_{t=1}^n \hat{v}(x^{tr}_t) x^{tr}_{ti} x^{tr}_{tj} - E[x^{te}_{i} x^{te}_J] \right| \geq M d_n \right) = o(1),   \\
        &P\left( \max\limits_{0 \leq i \leq p_n} \left| \frac{1}{n} \sum\limits_{t=1}^n \hat{v}(x^{tr}_t) x^{tr}_{ti} \varepsilon^{tr}_{t} \right| \geq M d_n \right) = o(1). 
    \end{align}
    These lead to the bounds:
    \begin{align}
        \begin{split}
            & P\left( \max_{0\leq i, j\leq p_n} \left|\frac{1}{n}\sum_{t=1}^n \hat{w}_{t} x^{tr}_{ti}x^{tr}_{tj} - E[x^{te}_ix^{te}_j] \right| > \overline{B} d_n \right) = o(1),\\
            & P\left( \max_{0\leq i \leq p_n} \left|\frac{1}{n}\sum_{t=1}^n \hat{w}_{t} x^{tr}_{ti}\varepsilon^{tr}_t \right| > \overline{B} d_n \right) = o(1),
        \end{split}
    \end{align}
    which in turn allow us to establish 
    Lemma \ref{lem:lem5} -- Lemma \ref{lem:lem11},
    with $\hat{w}_t$ replacing $w_t$,
    by applying arguments similar to those used in 
    Section \ref{sec:sec5}. In addition, \eqref{eq:eq9} and \eqref{imr20Mar18_1} yield 
    \begin{align}
        \frac{1}{n}\sum\limits_{t=1}^n \hat{w}_t(\varepsilon^{tr}_t)^2 = \sigma^2 + o_p(1), \label{eq:eq56}
    \end{align}
    ensuring that $\check{\sigma}^2(\hat{J}_{m_n}) \leq \overline{M}_1 + o_p(1)$ for some $\overline{M}_{1} > 0$. The proof of Theorem \ref{thm:thm3}
    can now be completed using arguments similar to those employed in the proof of  Theorem \ref{thm:thm2}.


\subsection{Proof of Theorem \ref{thm:thm5}}
\label{sec:sec7}
    We first define for $J \subset J_F$, 
    \begin{align}
        y^{tr}(\bm{x}|J)&:= \alpha^{tr}(J) + \bm{\beta}^{tr}(J)^\top \bm{x}_J,\\
        \varepsilon^{tr}(\bm{x}|J)&:= y(\bm{x}) - y^{tr}(\bm{x}|J) = (\bm{x} - \bm{\mu}^{tr})^\top (\bm{\beta} - \bm{\beta}^{tr\star}(J)),  \\
        \bm{\varepsilon}^{tr}(\bm{X}^{tr}|J)&:= (\varepsilon^{tr}(\bm{x}^{tr}_1|J), \ldots, \varepsilon^{tr}(\bm{x}^{tr}_n|J))^\top = (\bm{X}^{tr} - \bm{1}_n(\bm{\mu}^{tr})^\top) (\bm{\beta} - \bm{\beta}^{tr\star}(J)), 
    \end{align}
    where the latter two quantities
    are the training-data counterparts of
    $\varepsilon^{te}(\bm{x}|J)$ and $\bm{\varepsilon}^{te}(\bm{X}^{tr}|J)$. 
    Recall that $\hat{\alpha}^{tr}(J)$ and $\hat{\bm{\beta}}^{tr}(J)$ are $\hat{\alpha}(J)$ and $\hat{\bm{\beta}}(J)$ with $w_t = 1$; see
    \eqref{ing250019}. 
    %
    By treating the training distribution as the target distribution,
    \eqref{eq:eq6} and \eqref{eq:eq7} continue to hold with $\varepsilon^{te}(\bm{x}^{te}|\hat{J}_k)$ replaced by $\varepsilon^{tr}(\bm{x}^{tr}|\hat{J}_k)$. Concretely, we have
    \begin{align}
        \lim_{n \rightarrow \infty} P(\tilde{k}^{tr} \leq \hat{k} \leq V_\varepsilon  m_n^{tr}) = 1, 
        \label{eq:eq129}
    \end{align}
    where $m_n^{tr} = \lfloor c_n^{-\frac{1}{1 + \xi}}\rfloor - 1$,  
    \begin{align}
        \tilde{k}^{tr} := \min\left\{ k \in \{1, \ldots, K_n\} \mid E_n [\varepsilon^{tr}(\bm{x}^{tr}|\hat{J}_k)^2] \leq G_\varepsilon (m_n^{tr})^{-(1+2\xi)} \right\}, 
    \end{align}
    for some large but finite $G_\varepsilon$, and $\min \emptyset = K_n$. 

    It is clear that
    \begin{align}
    \begin{split}
        &E_n[(y(\bm{x}^{te}) - \hat{y}^{tr}(\bm{x}^{te}|\hat{J}_{\hat{k}}))^2 ] \\
        &\leq 2E_n[ (y(\bm{x}^{te}) - y^{tr}(\bm{x}^{te} | \hat{J}_{\hat{k}}) )^2 ]  \\
        &\phantom{\leq} + 2E_n [ \{ (\hat{\alpha}^{tr}(\hat{J}_{\hat{k}}) - \alpha^{tr}(\hat{J}_{\hat{k}}) ) +  ( \hat{\bm{\beta}}^{tr}(\hat{J}_{\hat{k}}) - \bm{\beta}^{tr}(\hat{J}_{\hat{k}}) )^\top \bm{x}^{te}_{\hat{J}_{\hat{k}}} \}^2 ], 
        \label{ing250030}
    \end{split}
    \end{align}
    where the first term is the bias, and the second term is the variance. 
	By (C\ref{cond:cond7}), we express
    $E_n[ (y(\bm{x}^{te}) - y^{tr}(\bm{x}^{te} | \hat{J}_{\hat{k}}) )^2 ]$ as:
	\begin{align}
   \begin{split}
        &E_n[ \{ (\bm{\beta} - \bm{\beta}^{tr\star}(\hat{J}_{\hat{k}}))^\top ( \bm{x}^{te} - \bm{\mu}^{te} ) - ( \bm{\beta}^{tr\star}(\hat{J}_{\hat{k}}) - \bm{\beta} )^\top (\bm{\mu}^{te} - \bm{\mu}^{tr}) \}^2 ]  \\
        &= ( \bm{\beta} - \bm{\beta}^{tr\star}(\hat{J}_{\hat{k}}) )^\top \bm{\Sigma}^{te} ( \bm{\beta} - \bm{\beta}^{tr\star}(\hat{J}_{\hat{k}}) ) + ((\bm{\beta}^{tr\star}(\hat{J}_{\hat{k}})  -  \bm{\beta})^\top (\bm{\mu}^{te} - \bm{\mu}^{tr}))^2.  \label{eq:eq130} 
    \end{split}
    \end{align}
Thus,
\begin{align}
   \begin{split}
    &E_n[ (y(\bm{x}^{te}) - y^{tr}(\bm{x}^{te} | \hat{J}_{\hat{k}}) )^2 ] \\
        &\leq \hat{k} \Vert \bm{\beta} - \bm{\beta}^{tr\star}(\hat{J}_{\hat{k}}) \Vert_2^2 \left( \max\limits_{1 \leq i \leq j \leq p_n}| \sigma^{te}_{ij}| + 2\Vert \bm{\mu}^{te}\Vert_{\infty}^2 + 2\Vert \bm{\mu}^{tr} \Vert_{\infty}^2 \right). 
        \label{eq:eq131}
	\end{split}
    \end{align}
 By \eqref{eq:eq62}, 
    \eqref{eq:eq1033} (applied to the special case where $f^{te}(\cdot)=f^{tr}(\cdot)$),
    and the definition of 
$\tilde{k}^{tr}$, we have
    \begin{align}
    \begin{split}
        \Vert \bm{\beta} - \bm{\beta}^{tr\star}(\hat{J}_{\hat{k}}) \Vert_2^2 I_{\{\tilde{k}^{tr} \leq \hat{k} \leq V_\varepsilon  m_n^{tr}\}} &\leq (C^{tr}_{\lambda+})^{-1} E_n[\varepsilon^{tr}(\bm{x}^{tr}|\hat{J}_{\hat{k}})^2]I_{\{\tilde{k}^{tr} \leq \hat{k} \leq V_\varepsilon  m_n^{tr}\}}\\
        &\leq (C^{tr}_{\lambda+})^{-1}G_\varepsilon (m_n^{tr})^{-(1+2\xi)},
        \label{eq:eq104}
        \end{split}
    \end{align}
where the second inequality corresponds
to (3.50) of \cite{ing2020model}.
Note, however, that 
 (3.50) in \cite{ing2020model}
contains an error: the first inequality should be removed, and the constant in the final bound should be replaced by 
$G$ as defined in  
(3.33) of that paper.
    On the other hand, (C\ref{cond:cond2}) and (C\ref{cond:cond8}) yield
    \begin{align}
        \max\limits_{1 \leq i \leq j \leq p_n}| \sigma^{te}_{ij}| + 2\Vert \bm{\mu}^{te}\Vert_{\infty}^2 + 2\Vert \bm{\mu}^{tr} \Vert_{\infty}^2 \leq C. 
        \label{imr250506-10}
    \end{align}
    Consequently, by \eqref{eq:eq129}, \eqref{eq:eq131},\eqref{eq:eq104} and \eqref{imr250506-10} ensure that, with high probability,
    \begin{align}
            E_n[ (y(\bm{x}^{te}) - y^{tr}(\bm{x}^{te} | \hat{J}_{\hat{k}}) )^2 ] &\leq C (C^{tr}_{\lambda+})^{-1}G_\varepsilon V_\varepsilon (m_n^{tr})^{-2\xi}. 
            \label{imr250506-9}
	\end{align}
    
    Next, we evaluate the variance term on the right-hand side of \eqref{ing250030}. 
    Arguing similarly to the proof of Lemma
    \ref{lem:lem11}, it follows that with high probability, 
    \begin{align}
        (\hat{\bm{\beta}}^{tr}(\hat{J}_k)^\top ( \bm{\mu}^{tr}_{\hat{J}_k} - \hat{\bm{\mu}}^{tr}_{\hat{J}_k} ) + \hat{\mu}^{tr}_y - \mu^{tr}_y  )^2 \leq \overline{B}c_n^2, \,\,\mbox{for all}\,\, k = 1, \ldots, K_n,
        \label{imr250509-1}
    \end{align}
    Moreover, similar to \eqref{imr250506-10}, it follows from (C\ref{cond:cond2}) and (C\ref{cond:cond8}) that with high probability,
    \begin{align}
        \max\limits_{1 \leq i \leq j \leq p_n}| \sigma^{tr}_{ij}| + 2\Vert \bm{\mu}^{te}\Vert_{\infty}^2 + 2\Vert \bm{\mu}^{tr} \Vert_{\infty}^2 \leq C. 
        \label{imr250509-2}
    \end{align}
    By \eqref{imr250509-1} and \eqref{imr250509-2}, we  have
    \begin{align}
     \begin{split}
        &E_n [ \{ (\hat{\alpha}^{tr}(\hat{J}_{\hat{k}}) - \alpha^{tr}(\hat{J}_{\hat{k}}) ) +  ( \hat{\bm{\beta}}^{tr}(\hat{J}_{\hat{k}}) - \bm{\beta}^{tr}(\hat{J}_{\hat{k}}) )^\top \bm{x}^{te}_{\hat{J}_{\hat{k}}} \}^2 ]  \\
        &= E_n [  \{
                \hat{\mu}^{tr}_{y} - \mu^{tr}_{y} + ( \hat{\bm{\mu}}^{tr}_{\hat{J}_{\hat{k}}} - \bm{\mu}^{tr}_{\hat{J}_{\hat{k}}} )^\top  \hat{\bm{\beta}}^{tr}(\hat{J}_{\hat{k}})\\
        &\phantom{= E_n \{} 
                + ( \bm{\mu}^{te}_{\hat{J}_{\hat{k}}} - \bm{\mu}^{tr}_{\hat{J}_{\hat{k}}} )^\top  (\hat{\bm{\beta}}^{tr}(\hat{J}_{\hat{k}}) - \bm{\bm{\beta}}^{tr}(\hat{J}_{\hat{k}}) )\\
        &\phantom{= E_n \{} 
                + ( \bm{x}^{te}_{\hat{J}_{\hat{k}}} - \bm{\mu}^{te}_{\hat{J}_{\hat{k}}} )^\top  (\hat{\bm{\beta}}^{tr}(\hat{J}_{\hat{k}}) - \bm{\bm{\beta}}^{tr}(\hat{J}_{\hat{k}}) )
            \}^2 ]  \\
        &\leq 3 \{
                \hat{\mu}^{tr}_{y} - \mu^{tr}_{y} + ( \hat{\bm{\mu}}^{tr}_{\hat{J}_{\hat{k}}} - \bm{\mu}^{tr}_{\hat{J}_{\hat{k}}} )^\top  \hat{\bm{\beta}}^{tr}(\hat{J}_{\hat{k}})
            \}^2  \\
        &\phantom{\leq} + 3
            \{
                ( \bm{\mu}^{te}_{\hat{J}_{\hat{k}}} - \bm{\mu}^{tr}_{\hat{J}_{\hat{k}}} )^\top  ( \hat{\bm{\beta}}^{tr}(\hat{J}_{\hat{k}}) - \bm{\bm{\beta}}^{tr}(\hat{J}_{\hat{k}}) )
            \}^2  \\
        &\phantom{\leq} + 3
            E_n [\{
                ( \bm{x}^{te}_{\hat{J}_{\hat{k}}} - \bm{\mu}^{te}_{\hat{J}_{\hat{k}}} )^\top  (\hat{\bm{\beta}}^{tr}(\hat{J}_{\hat{k}}) - \bm{\bm{\beta}}^{tr}(\hat{J}_{\hat{k}}) )
            \}^2]\\
        &\leq 3\overline{B}c_n ^2
            + 3 (\Vert \bm{\Sigma}^{te}_{\hat{J}_{\hat{k}},\hat{J}_{\hat{k}}} \Vert_2  + \Vert \bm{\mu}^{te}_{\hat{J}_{\hat{k}}} - \bm{\mu}^{tr}_{\hat{J}_{\hat{k}}} \Vert_2^2) \Vert \hat{\bm{\beta}}^{tr}(\hat{J}_{\hat{k}}) - \bm{\bm{\beta}}^{tr}(\hat{J}_{\hat{k}}) \Vert_2^2\\
        &\leq 3\overline{B}c_n ^2
            + 3 \hat{k} \left(\max\limits_{1 \leq i \leq j \leq p_n}|\sigma^{tr}_{ij}|  + 2\Vert \bm{\mu}^{te}\Vert_\infty^2 + 2\Vert \bm{\mu}^{tr}\Vert_\infty^2\right) \Vert \hat{\bm{\beta}}^{tr}(\hat{J}_{\hat{k}}) - \bm{\bm{\beta}}^{tr}(\hat{J}_{\hat{k}}) \Vert_2^2\\
        &\leq 3\overline{B}c_n ^2
            + 3 C \hat{k} \Vert \hat{\bm{\beta}}^{tr}(\hat{J}_{\hat{k}}) - \bm{\bm{\beta}}^{tr}(\hat{J}_{\hat{k}}) \Vert_2^2. 
        \label{eq:eq132}
    \end{split}
    \end{align}
    In addition, by 
    \eqref{eq:eq20}
    (applied to the special case where $f^{te}(\cdot)=f^{tr}(\cdot)$),
    we have with high probability
	\begin{align}
            \Vert \hat{\bm{\beta}}^{tr}(\hat{J}_k) - \bm{\beta}^{tr}(\hat{J}_k) \Vert_2^2 \leq \overline{B} kc_n^2, \,\,
            \mbox{for all}\,\, k = 1, \ldots, K_n. 
            \label{eq:eq120}
	\end{align}
    Hence, by \eqref{eq:eq129}, \eqref{eq:eq132}, and \eqref{eq:eq120}, with high probability, 
    \begin{align}
        &E_n [ \{ (\hat{\alpha}^{tr}(\hat{J}_{\hat{k}}) - \alpha^{tr}(\hat{J}_{\hat{k}}) ) +  ( \hat{\bm{\beta}}^{tr}(\hat{J}_{\hat{k}}) - \bm{\beta}^{tr}(\hat{J}_{\hat{k}}) )^\top \bm{x}^{te}_{\hat{J}_{\hat{k}}} \}^2 ] \\
        &\leq 3\overline{B}(1 + C)V_{\varepsilon}^2 (m_n^{tr})^2c_n^2. 
        \label{imr250505-3}
    \end{align}
    Combining \eqref{ing250030}, \eqref{imr250506-9} and \eqref{imr250505-3} with the definition of $m_n^{tr}$,
    we obtain \eqref{eq:eq95}. 

	To prove \eqref{eq:eq103}, note first that 
     \eqref{eq:eq92} and \eqref{eq:eq130} yield
    \begin{align}
    \begin{split}
        &E_n[(y(\bm{x}^{te}) - y^{tr}(\bm{x}^{te} | \hat{J}_{\hat{k}}))^2]  \\
        &\leq \Vert \bm{\beta} - \bm{\beta}^{tr\star}(\hat{J}_{\hat{k}}) \Vert_2^2 \left( \Vert \bm{\Sigma}^{te} - \bm{\Sigma}^{tr} \Vert_2 + \Vert \bm{\mu}^{te} - \bm{\mu}^{tr} \Vert_2^2 \right) + E_n[\varepsilon^{tr}(\bm{x}^{te}|\hat{J}_{\hat{k}})^2]  \\
        &\leq C_{\rm diff}(1 + C_{\rm diff})\Vert \bm{\beta} - \bm{\beta}^{tr\star}(\hat{J}_{\hat{k}}) \Vert_2^2 + E_n[\varepsilon^{tr}(\bm{x}^{te}|\hat{J}_{\hat{k}})^2]. 
        \label{eq:eq102}
        \end{split}
	\end{align}
    Since $E_n[\varepsilon^{tr}(\bm{x}^{te}|\hat{J}_k)^2]$ decreases as $k$ grows,
    it follows from 
    \eqref{eq:eq129}, \eqref{eq:eq104}, \eqref{eq:eq102},
    and the definition of $\tilde{k}^{tr}$ that 
    with high probability,
    \begin{align}
        E_n[(y(\bm{x}^{te}) - y^{tr}(\bm{x}^{te} | \hat{J}_{\hat{k}}))^2] \leq  G_\varepsilon(1 + C_{\rm diff}(1 + C_{\rm diff}) (C^{tr}_{\lambda+})^{-1} )(m_n^{tr})^{-(1+2\xi)}. 
        \label{imr250506-1}
	\end{align}
    On the other hand, by \eqref{eq:eq129} and \eqref{eq:eq120}, with high probability, 
    \begin{align}
    \begin{split}
        &E_n [ \{ (\hat{\alpha}^{tr}(\hat{J}_{\hat{k}}) - \alpha^{tr}(\hat{J}_{\hat{k}}) ) +  ( \hat{\bm{\beta}}^{tr}(\hat{J}_{\hat{k}}) - \bm{\beta}^{tr}(\hat{J}_{\hat{k}}) )^\top \bm{x}^{te}_{\hat{J}_{\hat{k}}} \}^2 ]\\
        &\leq 3\overline{B}c_n ^2
        + 3 ( \Vert \bm{\Sigma}^{te} - \bm{\Sigma}^{tr} \Vert_2  + \Vert \bm{\mu}^{te} - \bm{\mu}^{tr} \Vert_2^2) \Vert \hat{\bm{\beta}}^{tr}(\hat{J}_{\hat{k}}) - \bm{\bm{\beta}}^{tr}(\hat{J}_{\hat{k}}) \Vert_2^2\\
        &\phantom{\leq} + (\hat{\bm{\beta}}^{tr}(\hat{J}_{\hat{k}}) - \bm{\bm{\beta}}^{tr}(\hat{J}_{\hat{k}}))^\top \bm{\Sigma}^{tr}_{\hat{J}_{\hat{k}},\hat{J}_{\hat{k}}}(\hat{\bm{\beta}}^{tr}(\hat{J}_{\hat{k}}) - \bm{\bm{\beta}}^{tr}(\hat{J}_{\hat{k}}))\\
        &\leq 3(2 + C_{\rm diff}(1 + C_{\rm diff}))\overline{B} \hat{k} c_n^2\\
        &\leq 3(2 + C_{\rm diff}(1 + C_{\rm diff}))\overline{B} V_\varepsilon m_n^{tr} c_n^2, 
        \label{imr250506-2}
    \end{split}
    \end{align}
    where the second inequality follows from
    Lemma \ref{lem:lem10} (applied to 
    the special case $f^{te}(\cdot) = f^{tr}(\cdot)$). 
    By \eqref{imr250506-1}, \eqref{imr250506-2}, and the definition of $m_n^{tr}$, the proof of \eqref{eq:eq103} is complete.
	\hfill $\blacksquare$ \\
	


\newpage

\begin{thebibliography}{7}
\providecommand{\natexlab}[1]{#1}
\providecommand{\url}[1]{\texttt{#1}}
\expandafter\ifx\csname urlstyle\endcsname\relax
  \providecommand{\doi}[1]{doi: #1}\else
  \providecommand{\doi}{doi: \begingroup \urlstyle{rm}\Url}\fi

\bibitem[Bastani(2021)]{bastani2021predicting}
Hamsa Bastani.
\newblock Predicting with proxies: Transfer learning in high dimension.
\newblock \emph{Management Science}, 67\penalty0 (5):\penalty0 2964–2984, May 2021.
\newblock ISSN 1526-5501.
\newblock \doi{10.1287/mnsc.2020.3729}.
\newblock URL \url{http://dx.doi.org/10.1287/mnsc.2020.3729}.

\bibitem[Imori and Ing(2025)]{Imori202X}
Shinpei Imori and Ching-Kang Ing.
\newblock Prediction and variable selection in high-dimensional misspecified regression models under covariate shift.
\newblock Manuscript, 2025.

\bibitem[Ing(2020)]{ing2020model}
Ching-Kang Ing.
\newblock Model selection for high-dimensional linear regression with dependent observations.
\newblock \emph{The Annals of Statistics}, 48\penalty0 (4), August 2020.
\newblock ISSN 0090-5364.
\newblock \doi{10.1214/19-aos1872}.
\newblock URL \url{http://dx.doi.org/10.1214/19-AOS1872}.

\bibitem[Li et~al.(2021)Li, Cai, and Li]{li2020transfer}
Sai Li, T.~Tony Cai, and Hongzhe Li.
\newblock Transfer learning for high-dimensional linear regression: Prediction, estimation and minimax optimality.
\newblock \emph{Journal of the Royal Statistical Society Series B: Statistical Methodology}, 84\penalty0 (1):\penalty0 149–173, November 2021.
\newblock ISSN 1467-9868.
\newblock \doi{10.1111/rssb.12479}.
\newblock URL \url{http://dx.doi.org/10.1111/rssb.12479}.

\bibitem[Shimodaira(2000)]{shimodaira2000jspi}
Hidetoshi Shimodaira.
\newblock Improving predictive inference under covariate shift by weighting the log-likelihood function.
\newblock \emph{Journal of Statistical Planning and Inference}, 90\penalty0 (2):\penalty0 227--244, 2000.
\newblock ISSN 0378-3758.
\newblock \doi{https://doi.org/10.1016/S0378-3758(00)00115-4}.
\newblock URL \url{https://www.sciencedirect.com/science/article/pii/S0378375800001154}.

\bibitem[Sugiyama and Kawanabe(2012)]{sugiyama2012machine}
Masashi Sugiyama and Motoaki Kawanabe.
\newblock Applications of covariate shift adaptation.
\newblock In \emph{Machine Learning in Non-Stationary Environments}, pages 137--179. The MIT Press, March 2012.
\newblock \doi{10.7551/mitpress/9780262017091.003.0007}.
\newblock URL \url{http://dx.doi.org/10.7551/mitpress/9780262017091.003.0007}.

\bibitem[Tian and Feng(2022)]{tian2022transfer}
Ye~Tian and Yang Feng.
\newblock Transfer learning under high-dimensional generalized linear models.
\newblock \emph{Journal of the American Statistical Association}, 118\penalty0 (544):\penalty0 2684–2697, June 2022.
\newblock ISSN 1537-274X.
\newblock \doi{10.1080/01621459.2022.2071278}.
\newblock URL \url{http://dx.doi.org/10.1080/01621459.2022.2071278}.

\end{thebibliography}
\end{document}